\documentclass[journal]{IEEEtran}

\usepackage{amsmath,amsfonts}
\usepackage{algorithm}
\usepackage{array}
\usepackage[caption=false,font=normalsize,labelfont=sf,textfont=sf]{subfig}
\usepackage{textcomp}
\usepackage{stfloats}
\usepackage{url}
\usepackage{verbatim}
\usepackage{graphicx}
\usepackage{cite}
\usepackage{booktabs}
\usepackage{graphicx}
\usepackage{algpseudocode}
\usepackage{multirow}
\usepackage{array}
\usepackage[T1]{fontenc}
\usepackage{hyperref}

\usepackage[table,xcdraw]{xcolor}
\begin{document}

\title{RaidEnv: Exploring New Challenges in Automated Content Balancing for Boss Raid Games}

\newcommand{\orcidlogo}{\includegraphics[height=9pt]{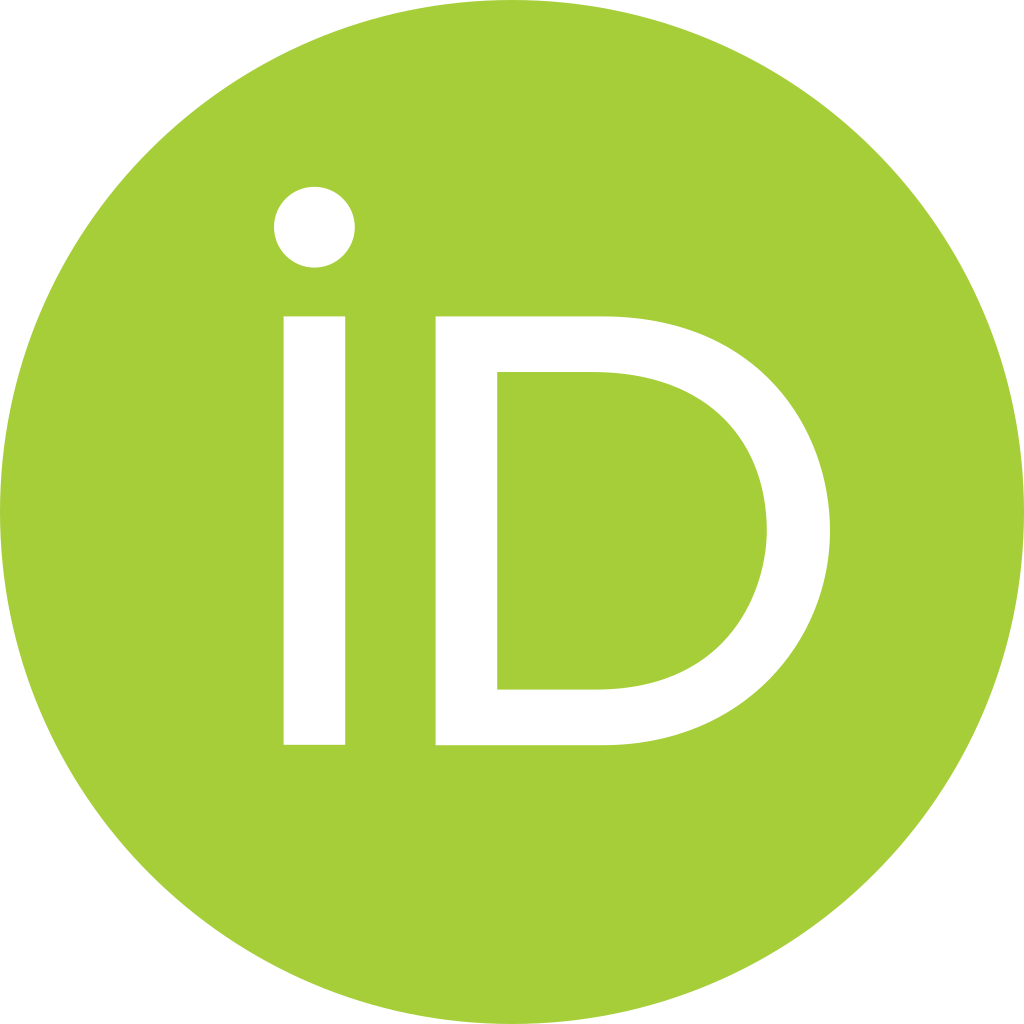}}

\newcommand{\orcidlink}[1]{\href{https://orcid.org/#1}{\orcidlogo}}

\author{Hyeon-Chang Jeon$^*$\orcidlink{0000-0003-3526-5191}, In-Chang Baek$^*$\orcidlink{0000-0002-9409-9253}, Cheong-mok Bae, Taehwa Park, Wonsang You, Taegwan Ha, Hoyun Jung, Jinha Noh, Seungwon Oh, Kyung-Joong Kim\orcidlink{0000-0002-7732-0817}

\thanks{equal contribution $*$}
\thanks{This paper was produced by the IEEE Publication Technology Group. They are in Piscataway, NJ.}
\thanks{Manuscript received April 19, 2021; revised August 16, 2021.}
}

\markboth{Journal of \LaTeX\ Class Files,~Vol.~14, No.~8, August~2021}%
{Shell \MakeLowercase{\textit{et al.}}: A Sample Article Using IEEEtran.cls for IEEE Journals}

\IEEEpubid{0000--0000/00\$00.00~\copyright~2021 IEEE}

\maketitle

\begin{abstract}
The balance of game content significantly impacts the gaming experience.
Unbalanced game content diminishes engagement or increases frustration because of repetitive failure.
Although game designers intend to adjust the difficulty of game content, this is a repetitive, labor-intensive, and challenging process, especially for commercial-level games with extensive content.
To address this issue, the game research community has explored automated game balancing using artificial intelligence (AI) techniques.
However, previous studies have focused on limited game content and did not consider the importance of the generalization ability of playtesting agents when encountering content changes.
In this study, we propose RaidEnv, a new game simulator that includes diverse and customizable content for the boss raid scenario in MMORPG games.
Additionally, we design two benchmarks for the boss raid scenario that can aid in the practical application of game AI.
These benchmarks address two open problems in automatic content balancing, and we introduce two evaluation metrics to provide guidance for AI in automatic content balancing.
This novel game research platform expands the frontiers of automatic game balancing problems and offers a framework within a realistic game production pipeline.
\end{abstract}

\begin{IEEEkeywords}
boss raid game environment, content generation, game playtesting, MMORPG
\end{IEEEkeywords}

\section{Introduction}

\IEEEPARstart{G}{ame} content balancing is a crucial process in the game industry when releasing new game content. Unbalanced game content can cause player dissatisfaction and frustration. To mitigate this issue, game companies employ balancing processes to prevent overpowered content that diminishes the game experience. Prior to content release, game testers evaluate the balancing and provide reports to the game designer, who verifies whether the content aligns with their intentions. Even after content release, game designers analyze game log data and readjust the content to achieve rebalancing.

However, several game companies rely on human testers repeatedly playing new game content to measure its difficulty. This approach has limitations as it does not allow testers to acquire sufficient data owing to time constraints. Moreover, the labor-intensive and expensive nature of this process limits the collection of extensive data from game testers. Recently, several scholars have attempted to employ machine learning techniques, such as in popular puzzle games \cite{poromaa2017crushing, gudmundsson2018human, lorenzo2020use} and card-based real-time strategy games \cite{liu2019playing}, for testing game content.

Automatic content balancing (ACB) is an automated technique that aims to readjust or recombine game content to ensure a balance of the game.
Within the game research community, ACB is regarded as a promising solution to address this challenge by leveraging AI player and generator (i.e., balancer) agents as game testers and designers, respectively. 
The ACB process involves two repetitive phases to automating game design tasks and quality assurance: (1) generating new game content using machine learning methods known as procedural content generation with machine learning (PCGML) \cite{summerville2018procedural}
In ACB, there are two repetitive phases to automating game design tasks and quality assurance: (1) generating new game content with machine learning (PCGML) \cite{summerville2018procedural} methods and (2) evaluating the content through playtesting with an AI player.
These automated sequences allow generator models to be trained through extensive trial-and-error iterations, surpassing the limited trials feasible with human testers. The quality of the generated content depends on the AI player’s robustness in playing diverse game contents and fairly evaluating them \cite{gisslen2021adversarial}. Similarly, the content generator should possess the ability to generate various contents according to designer-specified requirements.

In this paper, we present RaidEnv, a game environment that aims to encompass the features found in commercial-level MMORPG games, with a specific focus on the boss raid scenario, a popular content type in MMORPGs.
This highly customizable environment includes machine learning interfaces for playtesting and procedural content generation (PCG) agents. Our contribution includes the proposal of two benchmarks applicable to automatic content balancing (ACB) problems. These benchmarks consist of (1) training generalized playtesting agents and (2) training controllable content generation. We conduct the benchmarks within RaidEnv, providing deep reinforcement learning (DRL) methods as baselines, in addition to handwritten heuristics. Each benchmark was individually analyzed, and a combined analysis is presented at the end of this paper. The major contributions of this study are summarized as follows:

\begin{itemize}
    \item We developed RaidEnv, an integrated game simulator that supports extensive customization, facilitating content generation, and playtesting benchmarks.
    \item We propose a benchmark for evaluating playtesting agents, emphasizing their generalization ability when encountered with content variations.
    \newpage
    \item We propose a benchmark for skill content generation, focusing on controllability, and diversity aspects.
\end{itemize}

To introduce the new environment and benchmarks, we have organized the sections as follows: In Section \ref{sec:related_work}, wwe review previous studies to compare the features of our environment, covering existing game environments, and balancing scenarios.
Section \ref{sec:raidenv} provides a detailed description of the new game environment, outlining its key features, and components.
In Section \ref{sec:multiagent_benchmark}, we present our first benchmark, which focuses on the development of robust playtesting agents. To adequately evaluate the performance of the generated content, we proposed adjusted test performance metrics considering the varying difficulty with content evolution. 
In the following Section \ref{sec:pcg_benchmark}, we introduce the PCG benchamrk involves skill content generation and utilizes the playtesting agent proposed in the first benchmark. Using agent-based simulations, the utility of the PCG models was evaluated in terms of benchmarking controllability and diversity. Finally, the comprehensive results obtained from both benchmarks are discussed to highlight the potential implications of this research.

\begin{table*}[!ht]
\caption{Summary of the environments for contents generation and playtesting}
\center
\label{tab:env_summary}
\begin{tabular}{@{}lcccc@{}}
\toprule
\textbf{Environment} & \textbf{Ref.} & \textbf{PCG Problem} & \textbf{Multiagent Problem} & \textbf{Balancing Problem} \\ \midrule
Hearthstone & \cite{dockhorn2019introducing} & Card Deck Building  & - & - \\ \midrule
Ludii & \cite{stephenson2019ludii}  & Game Rule & - & - \\ \midrule

GVGAI & \cite{perez2019general} & Rule, Map Layout & - & - \\ \midrule
VizDoom & \cite{Wydmuch2019ViZdoom,Kempka2016ViZDoom} & \begin{tabular}[c]{@{}c@{}}Skill, Map Layout\end{tabular} & - & - \\ \midrule
Mario AI Framework & \cite{Ahmed2023GitHub,karakovskiy2012mario} & Map Layout & - & - \\ \midrule
Google Research Football & \cite{kurach2020google} & - & High-performance & - \\ \midrule
Fever Basketball & \cite{jia2020fever} & - & High-performance & - \\ \midrule
Heroes Charge & \cite{wang2021improved} & - & - & Single-content Balancing (Unit) \\ \midrule
Red Alert & \cite{preuss2018integrated} & - & - & Single-content Balancing (Unit) \\ \midrule
Sports Game & \cite{nikolakaki2020competitive} & - & - & Match Making \\ \midrule
Dungeon Replicants I \& II & \cite{pfau2020dungeons,pfau2022dungeons} & - & - & Experience-driven Playtesting \\ \midrule
SMAC I \& II & \cite{samvelyan2019starcraft,ellis2022smacv2} & Scenario & Environmental Generalization & - \\ \midrule
Melting Pot & \cite{leibo2021scalable} & Map Layout & Ad-hoc Generalization & - \\ \midrule
Neural MMO & \cite{suarez2019neural} & Massive Map Layout & High-performance & - \\ \midrule
Tower Defense & \cite{beau2016automated} & Map Layout & - & Multi-content Balancing (Spawn, Unit) \\ \midrule

Pokémon & \cite{reis2021vgc} & Card Deck Building & High-performance & Automatic Team Assembly \\ \midrule
\textbf{Our Environment} & \textbf{} & \textbf{Skill, Stats} & \textbf{\begin{tabular}[c]{@{}c@{}}Environmental Generalization\end{tabular}} & \textbf{Multi-content Balancing} \\ \bottomrule
\end{tabular}%
\end{table*}

\begin{figure*}[!ht]
    \centering
    \includegraphics[width=\linewidth]{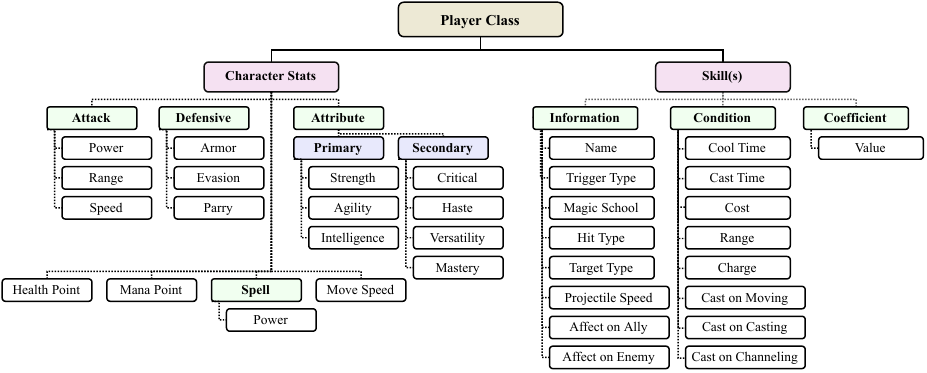}
    \caption{Overall game components in the RaidEnv environment. 
    The game features and environmental variables are illustrated in hierarchical structure and the same-level groups are colorized in same color (e.g., light red and light green).
    White box components are the modifiable variables and the colorized boxes are not modifiable, but grouped similar functional variables.
    \textbf{Character Stats} is the characteristic of the playtesting agents and the \textbf{Skill(s)} is ownable component by an agent. Each agent could have 1-3 executable skills and all the components are fully differentiate by the agent.}
    \label{fig:player_class_architecture}
\end{figure*}

\section{Related Works}
\label{sec:related_work}
We summarize the existing game research platforms in Table \ref{tab:env_summary} and the development purpose of them.
In Table \ref{tab:env_summary}, the balancing problem column indicates which problem was handled; the platforms that support content generation or multi-agent interfaces are additionally summarized for comparing the detailed features.
As shown in Table \ref{tab:env_summary}, the platforms have various problem definitions and only one (i.e., Pokémon) handled all three features: content generation, multiagent, and balancing.
That is, our platform has huge scalability for the game studies and propose new case study for novel problems.
\subsection{Automated Content Balancing}
In previous studies (refer to Balancing Problem in Table \ref{tab:env_summary}), four representative balancing problems have been proposed: automatic team assembly, matchmaking, experience-driven playtesting, and single-/multi-content balancing; The last problem is particularly relevant to our work.
Automatic team assembly aims to predict the power of a team \cite{gong2022automated} and identify overpowering champion combinations \cite{reis2021vgc} in battler games.
Matchmaking studies \cite{nikolakaki2020competitive} predict competitive combinations of two or more players, considering skill rating systems such as TrueSkill \cite{herbrich2006trueskill} and Elo \cite{arpad1978rating}).
Experience-driven playtesting studies aim to generate diverse behavioral playtesting logs from a game.
In \cite{pfau2020dungeons, pfau2022dungeons}, the authors explore a deep player behavior modeling approach, by training imitation models with the playtesting logs of 213 players.
Single- and multi-content balancing focuses on regulating game object parameters to ensure fair gameplay or intended symmetry.
Evolutionary methods are employed in \cite{preuss2018integrated,sorochan2022generating} to balance units (characters) in real-time games.
In particular, \cite{sorochan2022generating} simulates the game with playtesting agents of varying levels of proficiency.

Although the developed content generation task is similar to single-content balancing problems \cite{wang2021improved,preuss2018integrated}, it was applied in a different game genre, MMORPG.
Although our environment supports multi-content balancing for various game elements such as skills and stats, we demonstrated content generation for a single content to simplify the problem size.
Although this research shares certain similarities with the VGC AI competition \cite{reis2021vgc}, it exhibits notable differences.
First, the VGC AI competition focused on deck-building rather than altering the content itself, aiming to determine the possible combinations.
Second, they primarily focused on the framework, whereas our focus lies on evaluation.
Furthermore, this research emphasizes the robust playtesting of playtesting agents.
We propose a balancing study in a cooperative game scenario, specifically a boss raid that encompasses various game design components such as character class and executable skills.
We provide baseline approaches based on the latest DRL methods.
Notably, we highlight the significance of playtesting agents' generation ability in game-balancing studies.
Additionally, the proposed platform—utilizing a popular machine learning framework—have been published as an open-source model for game researchers to generate accessibility for multi-content generation tasks.
This platform presents numerous opportunities for exploring novel balancing scenarios among the game research community.
\subsection{Multiplayer Game AI Environments}
Multiplayer game AI environments primarily focus on designing algorithms that determine the behavior of each agent within a game system involving cooperation, competition, or both among two, or more players. Various environments have been proposed to facilitate multiplayer game AI studies, addressing specific problems using multiplayer game AI algorithms, as depicted in Table \ref{tab:env_summary}.
Several multiplayer game environments have concentrated on achieving superhuman-level AI performance in diverse games. 
For example, the Starcraft Multi-Agent Challenge (SMAC)\cite{samvelyan2019starcraft}, based on the RTS game StarCraft II, offers a range of unit configurations, emphasizing long-term decision-making, and partially observable contexts. 
In the sports game genre, Google Research Football (GRF)\cite{kurach2020google} and the Fever Basketball environment\cite{jia2020fever} aim to develop high-performing agents in sports-related games. 
Recently, the research focus has shifted toward environments that prioritize performance generalization, as observed in Melting Pot\cite{leibo2021scalable} and SMACv2\cite{ellis2022smacv2}. 
Melting Pot focuses on generalizing agent performance when  playing against previously unseen agents. It encompasses sequential social dilemmas (SSD), competitive games, and cooperative games.
SMACv2 aims to generalize agent performance in unseen environments by providing procedurally generated unit locations, types, and numbers.
Our study shares similarities with previous work\cite{ellis2022smacv2} in terms of the goal of achieving strong performance in unseen environments. However, our focus is on developing robust playtesting agents capable of adapting to a wider range of game contents. Additionally, our environment introduces varying levels of difficulty through changing game content.

\subsection{Procedural Contents Generation Environments}
Procedural content generation (PCG) studies have explored various types of game content generation, including game level layout, game rule generation, and card deck generation.
Game level layout generation has been extensively researched in popular video games such as Angry Birds \cite{kaidan2016procedural,ferreira2017tanager}, Super Mario Bros \cite{shu2021experience,volz2018evolving}, and Minecraft \cite{jiang2022learning}.
The objective of level generation is to create  playable game content that can be explored and completed by players.
Game rule generation involves automatically generating the mechanics, dynamics, and constraints of a game.
This area of research has been proposed in platforms such as general video game AI (GVGAI) \cite{perez2019general} and Ludii \cite{stephenson2019ludii}, utilizing game description languages.
Card deck generation focuses on searching for optimal combinations of game cards and has been studied in representative card simulation games such as Hearthstone \cite{dockhorn2019introducing} and Pokemon \cite{perez2019general}.
To expand the scope of PCG research to extensive game contents, new customizable game elements must be considered.
However, a limited number of studies have focused on generating game content such as skills and player characteristics.
Therefore, this study introduces a novel content type for procedural generation and demonstrates the generation 
 using reinforcement learning.
Game features such as skills, character classes, and items are common across various game genres, with numerous features shared within the same genre. 
To ensure the generality of our framework, we extract several features from commercial games and provide them as  controllable parameters.
This approach offers new perspectives on generating game content and opens up new possibilities in the game research community.

\section{RaidEnv: The Boss Raid Environment}
\label{sec:raidenv}
\subsection{Boss Raid Scenario}
\label{sec:bossraid_scenario}
Boss raids are a prevalent form of cooperative multiplayer content in most MMORPG games.
The objective of a raid is to defeat a powerful boss within a limited time frame in a dungeon, while playing alongside other players, with the aim of obtaining valuable rewards.
Boss raid content is centered around the challenge of overcoming a formidable boss with limited resources and time.
Typically, bosses are designed to be stronger than a single player, featuring complex attack patterns that pose a significant challenge in solo encounters.
Certain attack patterns necessitate player cooperation, such as spreading out to minimize damage when the boss launches an area attack. This highlights the importance of real-time communication between players.
As a fundamental principle in multiplayer game design \cite{seif2010understanding,zea2009design}, role differentiation is applied to boss raid content for assigning specific responsibilities to individual players.
In the boss raid scenario, role differentiation is achieved by diversifying the skills and abilities of players, with well-defined roles such as tankers, dealers, and healers being widely standardized.
Commercial MMORPG games often offer numerous character roles to enhance the cooperative gameplay experience. For instance, World of Warcraft (WoW)\footnote{https://worldofwarcraft.blizzard.com/} features over 30 specialized roles.

The increasing complexity of role types and game content poses challenges for game designers in predicting the outcomes of game updates and managing the combined effects of various elements. Therefore, the use of automatic content balancing (ACB) techniques becomes crucial for assisting game designers and developers in this complex landscape. However, only a few studies have attempted to apply ACB in the context of boss raids, and existing attempts considered an inadequate number of  customizable features to facilitate comprehensive balancing studies.
Consequently, the RaidEnv simulator offers a valuable contribution by providing a customization interface that enables the generation of diverse character roles through game simulation.
The implementation details of the boss raid scenario are described in the following section.
\subsection{Contents Definition}
\label{sec:content_definition}
To abstract boss raid content, we have categorized the game elements into three components: player class, character statistics, and skill parameters.
Figure \ref{fig:player_class_architecture} provides an overview of the design process for player classes, and the specific details of each component are described below:

\textbf{Class}: The player class is defining characteristic that imparts individuality to player characters and assigns the distict roles within the game.
The class of a character consists of two key elements: statistics and a set of skills.
For example, the Tanker class specializes in withstanding enemy attacks, possessing unique statistics such as high health points and defensive bonuses compared to other classes.
In addition, tankers possess skills that enable them to mitigate incoming damage and protect their allies.

\textbf{Statistics (Stats)}: Statistics represent the internal attributes of characters in RPGs and are typically represented as numerical values.
RPGs, ranging from traditional tabletop games to modern MMORPGs, utilize various statistics to define character attributes.
These statistics can be gained permanently or temporarily through character growth, equipment, consumables, or special abilities. Examples of statistics include health/mana points, strength, intelligence, dexterity, critical chances, and more.

\textbf{Skill}: While modern games employ different terms such as abilities, talents, or traits, in our environment, we simplify these concepts into the common notion of skills.
Skills are possessed by characters and can be categorized into two types: active skills and passive skills.
Active skills are initiated by a character's action, allowing the player to enact decisions regarding which skill to activate from their available set of skills.
Conversely, passive skills are latent abilities that have an impact even when not directly activated by the character.
Skills can be differentiated by analyzing conditions for activation, target affected, effects produced, and other relevant factors.

The environment offers numerous customization features, including 17 character statistics, and 17 player skill parameters.
These variables can be included or excluded as needed, providing a highly versatile game research platform.
For the PCG track in this study, we have chosen four representative skill parameters, as generating all skill parameters simultaneously would introduce a complex problem size.
However, the environment supports customizable variable across multiple game components, allowing for future exploration of multi-content balancing.

\subsection{RaidEnv: The Game Simulation Environment}
\begin{figure}[!h]
    \centering
    \includegraphics[width=\linewidth]{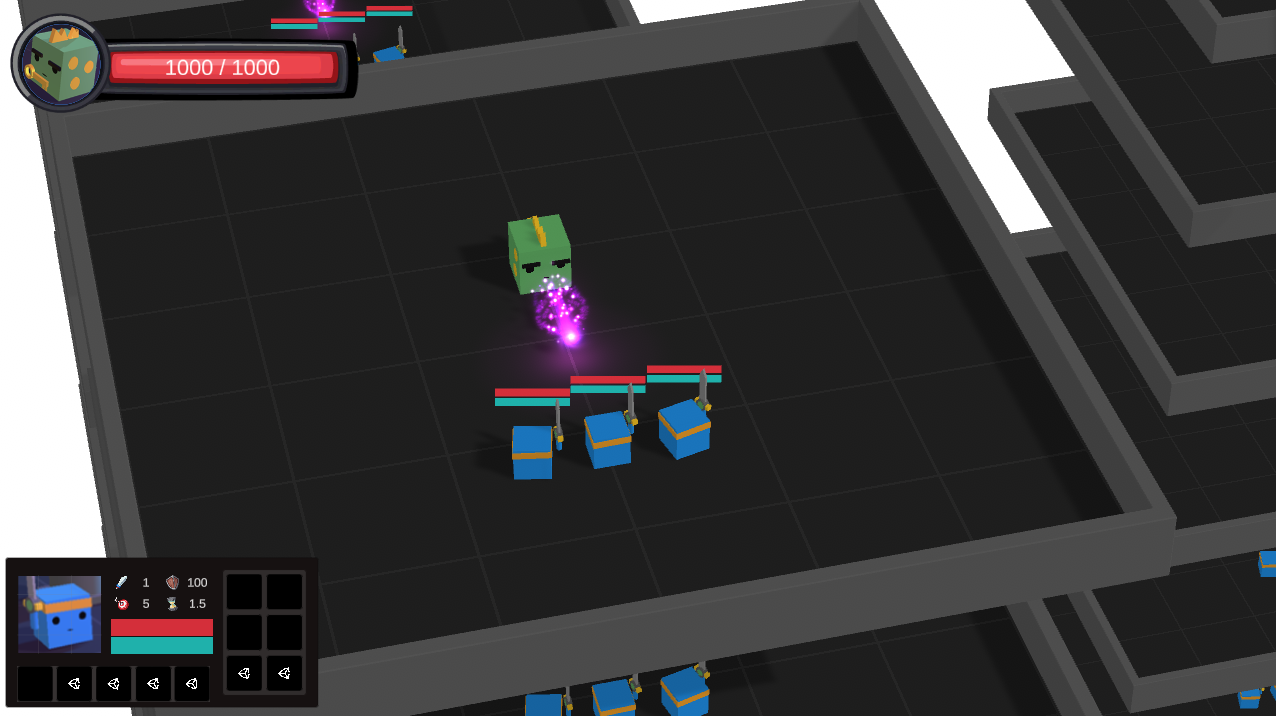}
    \caption{The snapshot of the RaidEnv environment. The blue ones are the player agents and the green one is the boss agent.}
    \label{fig:raidenv_snapshot}
\end{figure}

\label{sec:unity_mlagents}
The RaidEnv environment is implemented using the Unity platform, and for machine learning training, we utilized Unity ML-Agents Toolkit \cite{juliani2020}.
With the RaidEnv environment as a foundation, we have designed several scenarios tailored to each benchmark. 
During each training step, the single unity agent environment receives continuous observations (e.g., scalars, ray perception sensors) and samples discrete actions for learning within the environment. The RaidEnv environment encompasses a basic MMORPG boss raid scenario, consisting of one enemy agent and three player agents tasked with defeating the boss. Each player agent possesses one skill targeted at the enemy.
To efficiently gather playtesting results during training, parallelization of the environment is necessary. Therefore, we have included multiple arenas within the environment to enable parallel playtesting. 
Furthermore, We created interfaces to promote learning among multiplayer agents and PCG, including functions to compile game logs for tracing action steps and episodes.

\begin{figure*}[!ht]
    \centering
    \includegraphics[width=\linewidth]{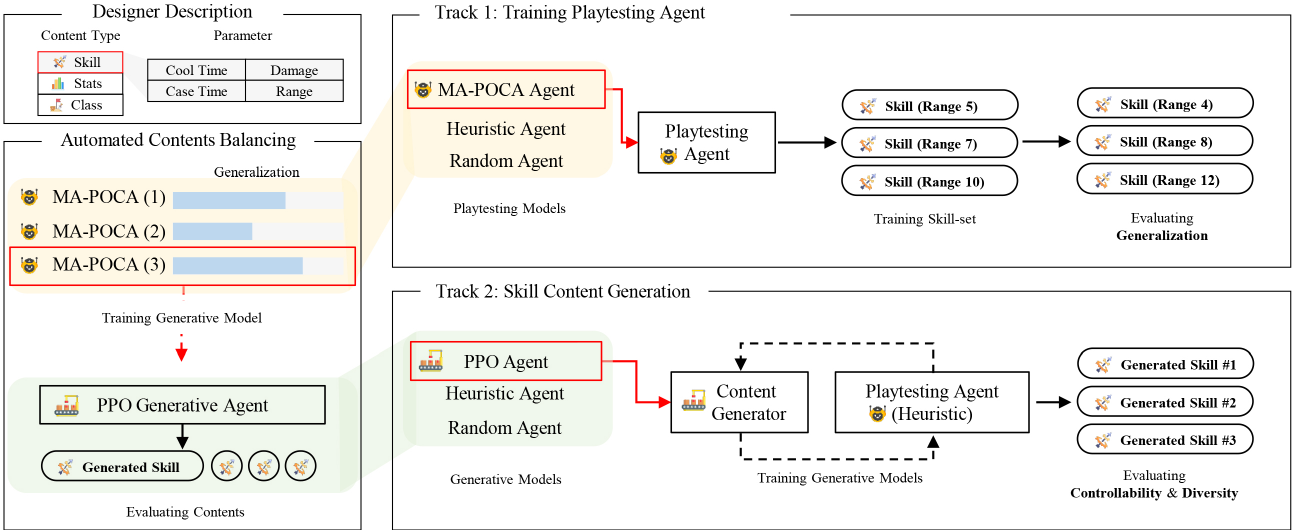}
    \caption{The proposed two benchmarks in the RaidEnv. The overall balancing process includes training playtesting agents and content generation with the playtesting agents process. In this work, we separated the two process as benchmarks to clarify the benchmark tasks. The integrated process (Automated Contents Balancing) is considered as the future work by integrating the results from two benchmarks.}
    \label{fig:benchmark_tracks}
\end{figure*}

The \textbf{Boss} in the boss raid scenario serves as the opponent non-player character (NPC).
As mentioned in Section \ref{sec:bossraid_scenario}, the boss is intentionally designed to be more powerful than a single player, fostering cooperation among the players.
The boss has 10 times the health points of a player agent and is equipped with two strong skills with ranges of 6 and 12.
The boss agent follows simple behavioral policy: it selects the closest player agent as its target, moves toward that player, and launches an attack when the skill is available. Figure \ref{fig:raidenv_snapshot} provides a snapshot of the environment, featuring one boss NPC, and three player agents
In summary, the RaidEnv environment consists of one static boss NPC and three customizable player agents. Two benchmarks are proposed within this setting, and the overall process is depicted in Figure \ref{fig:benchmark_tracks}. We will first delve into the playtesting aspect in Section \ref{sec:multiagent_benchmark}, followed by a discussion on content generation in Section \ref{sec:pcg_benchmark}.


\section{Benchmark 1: Playtesting Agent with Changing contents}
\label{sec:multiagent_benchmark}
In this section, we present a playtesting agent trained using multiagent reinforcement learning (MARL) and evaluate its content robustness to different game contents.
To effectively utilize the playtesting agent in various scenarios, the agent should manifest adaptability to varing content.
Therefore, we assess whether the agent maintains consistent performance across different skill parameter value settings within the environment.

Furthermore, to reflect the variation in difficulty resulting from content variations, we propose the use of a generalization score metric that considers difficulty adjustments.
\subsection{Environment Settings for Playtesting Agent}
\label{sec:roub_rep}
\textbf{Agent}
In this section, we introduce the MARL agent, MA-POCA\cite{cohen2021use}.
MA-POCA addresses the posthumous credit assignment problems, which arises when an agent is removed from the environment during training.
This problem is particularly critical in games where agent death is a common occurrence.
Specifically, MA-POCA is constructed using the counterfactual multi-agent policy gradient (COMA)\cite{foerster2018counterfactual}, incorporating the self-attention layer\cite{vaswani2017attention} in the network to handle posthumous credit assignment problems.
The objective of MA-POCA involves utilizing a counterfactual baseline\cite{wolpert2001optimal} and updating for agent $i$ is performed as follows:
\begin{align}
    \label{eq:ma-poca}
    \mathcal{J}^i_{critic} = \tau^i - \sum_aQ(s,(a^{-i}, a)) \\
    \mathcal{J}^i_{actor} = log\pi_i(a^i|s)Q(s, (a^{-i}, a^i))
\end{align}
where $\tau^i $ represents the temporal difference target, $Q$ denotes the state-value function, and $\pi_i$ represents the policy of agent $i$, applicable to both homogeneous, and heterogeneous agent settings.

\textbf{State}
At each time step, an agent collects observations on the boss and the status of all team members, including position, velocity, health, and remaining skill cooldown.
Additionally, the agent gathers information on existing skill projectiles.
Furthermore, to facilitate generalized behavior across skill parameters, the agent obtains parameters such as range, damage, and cast time associated with the skills. 
All features are normalized based on their maximum values.

\textbf{Action}
Agents have various actions, including move, rotate, and execute skills.
In our setup, each agent can execute a single action per time step, such as stay, move forward, move backward, turn right, turn left, move left, move right, and execute skill.

\textbf{Reward}
Agents receive individual rewards from the environment based on the amount of damage they inflict on the boss.
The damage reward is calculated as $ damage \times 0.01 $. 
Additionally, agents receive group rewards from the environment, which are set to $1.0$ when the boss is defeated.
Moreover, we introduce a cooperative element by incorporating a back attack reward.
A back attack occurs when one agent distracts the boss while other agents turn around and attack the boss' vulnerable rearside.
The reward for successful back attack is calculated as $ damage \times 0.012 $, which is higher than the reward for regular attacks.

\subsection{Playtesting Agent Evaluation Setup}
\label{sec:rob_exp_setup}
\subsubsection{Baselines}
RaidEnv requires a playtesting agent that can handle the complexity of MMORPG games while demonstrating robustness to content variations for PCG playtesting.
We employ three playtesting agents to validate the robustness of content variations: the reinforcement learning agent utilizing the MA-POCA algorithm\cite{cohen2021use}, the playtesting-heuristic (PT-HR) agent, and the playtesting-random (PT-RD) agent.

\begin{itemize}
    \item \textbf{MA-POCA} agent makes a decision about actions to defeat the Boss and receives rewards based on combat performance.
    \item \textbf{PT-HR} agent follows a simple strategy: it moves around the boss agent and randomly selects available skills during the combat.
    The agent maintains a maximum attack range from the boss and avoids its attacks.
    This is the baseline for comparing the multi-agent algorithm. Implementation details are provided in Appendix A.\ref{alg:Heuristic play-testing}.
    \item \textbf{PT-RD} agent selects actions from the action space by uniformly sampling distributions.
\end{itemize}

\subsubsection{Experiment Setup}
In this benchmark, we conducted two experiments: comparing the heuristic playtesting agent to the MARL agent using fixed content parameter settings and evaluating the agents` robustness. In the first experiment, we compare the basic performance of the PT-RD, PT-HR, and MA-POCA algorithms across various content parameter values, specifically the win rate for skill ranges of 5, 9, 13, and 17.
In the second experiment, we validate the agents’ robustness by varying the number of content parameter values, which reflects the diversity of the environment \cite{mckee2022quantifying}. At the beginning of each episode, the content parameter value is sampled from the environment, played with the sampled parameters, and used for learning about changing content parameters. In this experiment, The content parameters are set as skill ranges, with each agent having 1, 2, 3, and 4 skills. MA-POCA1 agent is sampled from skill ranges 5,  MA-POCA2 agent is sampled from skill ranges 5 and 9, MA-POCA3 agent is sampled from skill ranges 5, 9, and 13, while MA-POCA4 is sampled from skill ranges 5, 9, 13, and 17.

To evaluate the change of difficulty fairly, To evaluate the change of difficulty fairly, we made the population by extracting representative scores for a given level. The difficulty of level can be determined using various game logs such as churn rate\cite{roohi2020predicting,roohi2021predicting}, success rate\cite{gudmundsson2018human,kristensen2020estimating}. Herein, we evaluate the level of difficulty based on the win rate of games.

\subsubsection{Evaluation Metrics}

Utilizing the approach \cite{mckee2022quantifying} used to measure the generalization of performance on environment diversity, we measured the test gap between the heuristic playtesting agent and the MA-POCA agents with various skill range parameter values. We measured the performance of a setting that the trained skill range parameter settings contain in common, while the contents parameter setting that any agent has not seen was selected for the test.

Unlike the approach in \cite{mckee2022quantifying}, where the game difficulty remained constant across parameter changes, in our case varied with parameter changes. Therefore, these differences must be considered in-game difficulty. For example, the scenario becomes more challenging if the attack range of the skill is reduced by adjusting its range parameters, and we need to account for this difference. We used the term $score^{unseen}$ to denote the agent`s performance in the unseen contents parameter setting. 
Thereafter, We generated a population (N = 5) by training with the unseen contents parameter setting and obtained the average performance $score^{unseen}_{population}$ of the population. We then calculated the adjusted score $AdjustedScore$ using the following formula: 

\begin{align}
    \label{eq1:Train-Test Gap}
    AdjustedScore = \frac{score^{unseen}}{score^{unseen}_{population}}
\end{align}
In this metric, we defined the score as the win rate which represents explicit observable results and We make the MA-POCA populations in skill ranges 3, 4, 19, and 20, using range parameter values.

\subsection{Experimental Result \& Discussion}
\label{sec:rob_exp_result_discuss}

\subsubsection{MARL vs Heuristic}
\begin{table}[ht!]
\centering
\caption{Win rate between MARL and Heuristic agent}
\label{tab:robust_hr_rl}
\begin{tabular}{c|cccc@{}}
\toprule
Parameter &  \multicolumn{4}{c}{$Range$} \\ \midrule
Value & 5 & 9 & 13 & 17 \\ \midrule
MA-POCA & \begin{tabular}[x]{@{}c@{}}\textbf{0.485}\\($\pm{0.013}$)\end{tabular} & \begin{tabular}[x]{@{}c@{}}\textbf{0.857}\\($\pm{0.143}$)\end{tabular} & \begin{tabular}[x]{@{}c@{}}\textbf{0.990}\\($\pm{0.006}$)\end{tabular} & \begin{tabular}[x]{@{}c@{}}\textbf{0.990}\\($\pm{0.008}$)\end{tabular} \\
PT-Heuristic & \begin{tabular}[x]{@{}c@{}}0.089\\($\pm{0.017}$)\end{tabular} & \begin{tabular}[x]{@{}c@{}}0.371\\($\pm{0.025}$)\end{tabular} & \begin{tabular}[x]{@{}c@{}}0.698\\($\pm{0.021}$)\end{tabular} & \begin{tabular}[x]{@{}c@{}}0.830\\($\pm{0.089}$)\end{tabular} \\
PT-Random & \begin{tabular}[x]{@{}c@{}}0.000\\($\pm{0.000}$)\end{tabular} & \begin{tabular}[x]{@{}c@{}}0.002\\($\pm{0.002}$)\end{tabular} & \begin{tabular}[x]{@{}c@{}}0.029\\($\pm{0.006}$)\end{tabular} & \begin{tabular}[x]{@{}c@{}}0.058\\($\pm{0.013}$)\end{tabular} \\
\bottomrule
\end{tabular}
\end{table}

In this section, we present a comparison of the average win rate among the MA-POCA agent, the playtesting heuristic agent (PT-HR), and the playtesting random agent (PT-RD) over 500 games. The objective is to demonstrate why RL agents can be used to obtain a more reliable test result. The MA-POCA agent trained 50 million steps  for every 5 runs, according to the skill range setting. As in Table \ref{tab:robust_hr_rl}, MA-POCA agents both showed better results than PT-HR and PT-RD in the corresponding settings regardless of the range. In detail, the PT-HR agent shows a worse result as the skill range is decreasing. This indicates that the PT-HR has limitations on solving difficult levels of the game, while complex MA-POCA agent solves it well. The PT-RD agent cannot win even one game in range 5 because the game is much more difficult because the agents should get close to the boss. On the contrary, as range is increasing, even simple PT-RD agents can win the boss, which indicates the game is really easy.
The MA-POCA agent converged successfully, unlike the relatively fast convergence of skill ranges 13 and 17 settings, skill ranges 5 and 9 settings require more steps in convergence on training(See Table \ref{tab:robust_train_test_gap} in Appendix F).

\subsubsection{Robustness of MARL Agent}
 
\begin{figure}[!h]
    \centering
    \includegraphics[width=1.0\linewidth]{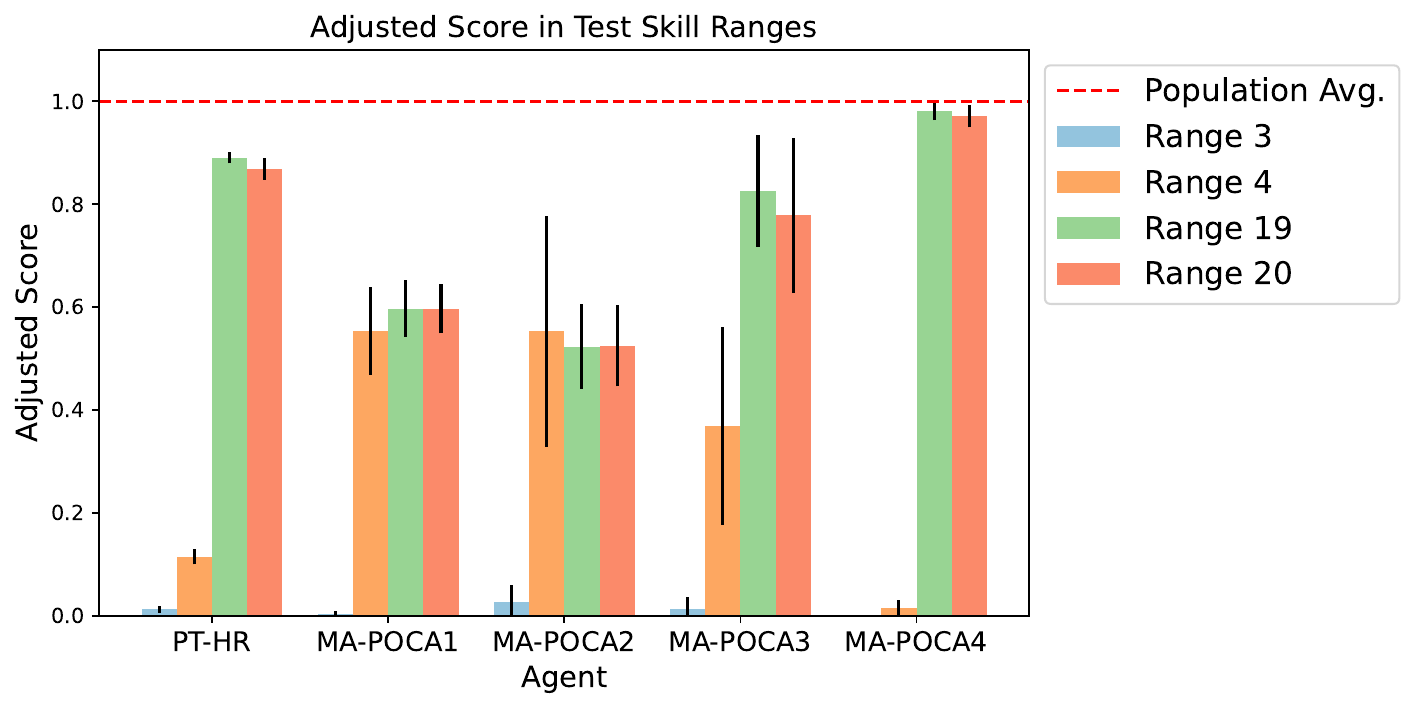}
    \caption{Figure of adjusted score results in test skill range settings. The red line indicates the average adjusted score of the population (N=5) in unseen settings. Adjusted score is the ratio of the test performance of the trained agent to the average performance of the population. As the adjusted score achieves a higher test adjusted score, the score is near 1 (sometimes even higher than 1).}
    \label{fig:barchart}
\end{figure}
We conducted each MA-POCA1, MA-POCA2, MA-POCA3, and MA-POCA4 5 runs training 50 million steps per set, which is based on the fastest convergence achieved by the MA-POCA1 agent. After the training, the performance on skill ranges 3, 4, 19, and 20, which were unseen during training were evaluated and compared. Figure \ref{fig:barchart}, illustrates the comparison between the trained agents and non-learning PT-HR agents for each MA-POCA agent. These figures depict the relative game difficulty achieved compared to the average performance of the population (N = 5), which serves as an indicator of the game difficulty at each skill range. In this case, the population was created by training separate MA-POCA agents for each range.

Overall, MA-POCA agents outperformed heuristic agents in the test content parameter settings. Among the MA-POCA agents, MA-POCA3 exhibited the highest average performance for the test content parameter settings (3, 4, 19, and 20) (refer to Table \ref{tab:robust_train_test_gap} in Appendix F). This indicates successful learning of strategies for different situations such as the boss’ melee range, between the boss’ melee and distant skills, and beyond the boss’ melee range. In contrast, MA-POCA4 displayed intensive learning of long-distance skills but struggled with short-distance encounters. MA-POCA1 demonstrated consistent performance across all test content parameter settings because of its reliance on a single strategy. Notably, all agents struggled to learn in the range 3 setting, suggesting room for further improvement and development.

\subsubsection{Discussion}
\begin{figure}[!h]
    \centering
    \includegraphics[width=1.0\linewidth]{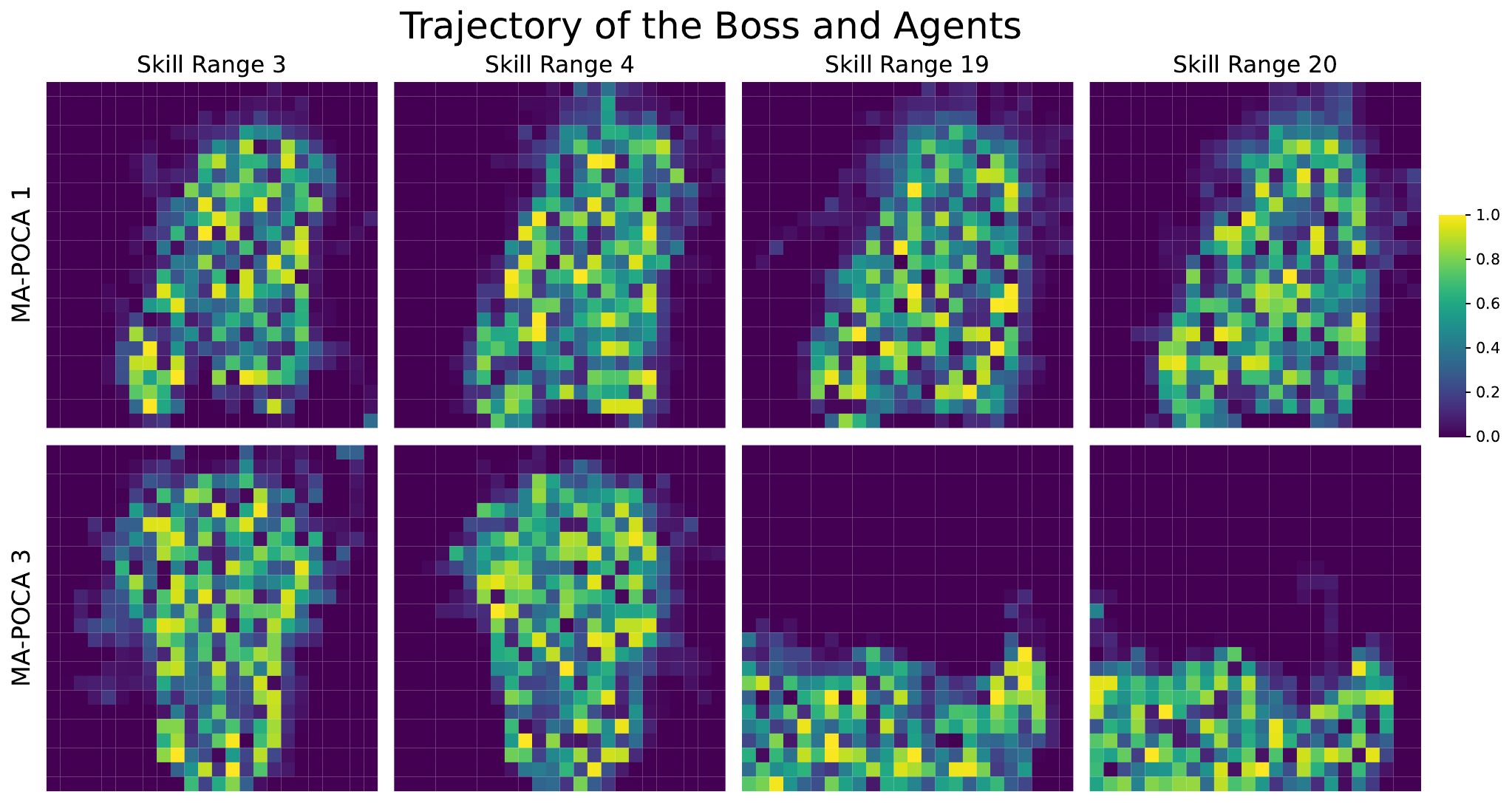}
    \caption{Heatmap of the average game position occupancy of MA-POCA1, MA-POCA3 in 500 games. The results show that MA-POCA1 agent does not change strategy well even skill range changes. On the contrary, MA-POCA3 agent recognized the change of parameter so that change the strategy of movement.}
    \label{fig:heatmap}
\end{figure}
We analyzed the movement patterns of MA-POCA1 and MA-POCA3 agents based on the number of sampled skill ranges. We played 500 games for each agent and averaged their occupied locations. As depicted in Figure \ref{fig:heatmap}, the MA-POCA1 agent, which only observed a skill range of 5, exhibited consistent behavior regardless of skill range changes. In contrast, the MA-POCA3 agent recognized the variation in skill ranges and adapted different strategies accordingly.

Although MARL algorithms have demonstrated successful learning across all content difficulties, a fully robust agent has yet to be achieved. Specifically, MA-POCA4 encountered difficulties in learning even with seen contents (range 5) (refer to Table \ref{tab:robust_train_test_gap} in Appendix F). This suggests that the MA-POCA4 agent struggled to adapt to challenging environments due to overfitting caused by its ease in learning from simpler environments. Furthermore, in the case of the content parameter range 3, representing extremely challenging levels, the agent failed to generalize altogether. Nevertheless, the MA-POCA agent showcases its strength in slightly more challenging environments. We anticipate that learning can be accelerated through approaches such as curriculum learning.

\section{Benchmark 2: Contents Generation}
\label{sec:pcg_benchmark}

\label{sec:pcg_descrition}
In this section, we present three baselines for the content generation task in the boss raid scenario.
The primary objective of designing procedural content generation (PCG) agents is to generate or rebalance game content to achieve specific target game outcomes.
For instance, a game designer may request the generator agent to produce a skill that yields a desired win rate (e.g., 60\%), and the agent should adjust the skill parameters to achieve the designated win rate.
This approach can be applied to both generating new game content from scratch and rebalancing existing content.
We introduce two notations for this task: (1) current win rate, $W_{c}$, which represents the simulated result with arbitrary game content, and (2) target win rate, $W_{t}$, which is the desired outcome of the generation process.
The ultimate goal of this task is to minimize the difference between $W_{c}$ and $W_{t}$ by adjusting the skill parameter values.

As an example of skill content generation, we propose a deep reinforcement learning (DRL) method. The DRL framework for PCG was initially introduced in \cite{khalifa2020pcgrl}, and the authors of \cite{earle2021learning} further enhanced this framework to improve its controllability.
As a machine learning approach for content generation, we adopt a DRL-based baseline utilizing the controllable PCGRL framework presented in the previous study \cite{khalifa2020pcgrl}.
The specific details of the DRL implementation for our work are described below.
To evaluate the generated content, we employ the heuristic playtesting agent discussed in the previous section (refer to Appendix A.\ref{alg:Heuristic play-testing}), and we refer to this agent as the playtesting heuristic (PT-HR) agent to avoid confusion with the heuristic PCG agent.

\subsection{Environment Settings for DRL}
\label{sec:representation}
\textbf{Agent}:
PCGRL (PCG via RL) \cite{khalifa2020pcgrl} proposed a new framework for generating game content by training a DRL-based generative model.
The state is defined as the scalar or tabular representation of game content; the action involves modifying the content; and the reward is measured based on the completeness of the generated content.
Subsequent works, such as controllable PCGRL have been introduced for two-dimensional \cite{earle2021learning} and three-dimensional games \cite{jiang2022learning} by designing goal-oriented reward functions that calculate the distance to a designer-defined value.
Controllable PCGRL enables the training of a generative model to produce a desired output specified by the designer, therby enhancing the usability of PCGRL models. 
In this study, we utilize controllable PCGRL to demonstrate the skill generation task in Section \ref{sec:pcg_benchmark}, training the generative model to generate diverse game outcomes.

\textbf{State}:
The state for the generator includes the skill parameter values required for generating the target skill.
The generator agent receives four scaled values representing the balancing parameters: cool time, range, damage, and cast time.
In our setup, each value is scaled between a minimum and maximum value. For example, the cool time, range is defined as 0.5-0.6, the skill range is set from 1 to 20, the damage range is between 0.5 and 1, and the cast time falls within the interval of 0.5-1.5.
For a comprehensive list of all environmental variables, please refer to Table \ref{tab:status_parameter} in Appendix \ref{sec:full_env_desc}.

\textbf{Action}:
The generator has an action space of $N$, where each action consists of five discrete values. Here, $N$ represents the number of skills, and the action determines the parameter updates.
We consider four parameters for skill generation ($N$ = 4): cool time, range, damage, and cast time.
The action determines the increment or decrement of each parameter using a hand-crafted scaler.
The granularity of the scaler was set to 0.16\% of each parameter range allowing for decrease (-) or increases (+) of [-0.16\%, -0.08\%, 0.0\%, +0.08\%, +0.16\%] within each parameter range.

\textbf{Reward}:
The reward function is designed to provide non-sparse reward signals to the generator. The generator receives positive rewards when it updates the skill parameters to achieve the target game win rate ($g$, goal).
To design the reward function for a controllable generator, we adopt a well-designed reward system proposed in the PCGRL framework \cite{earle2021learning,jiang2022learning}.
The L1 norm distance to the goal, $l$, is calculated as $l_t=abs(||g_t-p_t||_{L1})$, where $p_t$ represents the playtested win rate at time $t$. 
The reward at time $t$ is then calculated as $r_{t}=l_{t-1}-l_{t}$. A positive reward value indicates that the updated skill is closer to the target win rate.
In summary, the generator was trained to adjust the skill parameters for achieving the designer-defined playtesting result.

Notably, $p$ is measured by simulating the generated skills multiple times under the same conditions.
In our setup, we repeat the simulation ($N$ = 100) using the playtesting-heuristic agent (PT-HR) described in Section \ref{sec:rob_exp_setup}, and we report the average value of $p$ in the experimental results. The use of the PT-HR agent is preferred in PCG studies due to its reasonable performance and lower computational cost.

\subsection{Generator Evaluation Setup}
\label{sec:pcg_exp_setup}
\subsubsection{Baselines}
We compare the performance of three content generator models in the content generation experiment: the proposed PCG agent utilizing the proximal policy optimization (PPO) algorithm \cite{schulman2017proximal}, PCG-heuristic (PCG-HR) agent, and PCG-random (PCG-RD) agent.
PPO is a popular DRL algorithm known for its stability and has been widely used in previous content generation studies \cite{khalifa2020pcgrl,jiang2022learning}.
To provide performance comparisons for the PPO agent, we include a simple heuristic agent and a random agent.

\begin{itemize}
    \item \textbf{PPO} agent is based on the DRL approach and is implemented using the PCGRL framework, as detailed in Section \ref{sec:representation}. PPO agent enacts decisions to update skill parameters and receives rewards based on the playtesting results.
    \item \textbf{PCG-HR} agent randomly adjusts one of the four skill parameters by comparing the playtesting and target win rates. Specifically, The agent increases or decreases the parameter value according to a predefined rule based on human knowledge. The details of the implementation can be found in Algorithm \ref{alg:heuristic_pcg} in Appendix \ref{sec:pcg-hr algo}.
    \item \textbf{PCG-RD} agent adjusts the parameters without considering the target win rate. The agent randomly samples one action from five possible action branches in uniform probability. Each random action is applied to the four skill parameters.
\end{itemize}

\makeatletter
\def\algbackskip{\hskip-\ALG@thistlm}
\makeatother

\algnewcommand\algorithmicswitch{\textbf{switch}}
\algnewcommand\algorithmiccase{\textbf{case}}
\algnewcommand\algorithmicassert{\texttt{assert}}
\algnewcommand\Assert[1]{\State \algorithmicassert(#1)}%
\algdef{SE}[SWITCH]{Switch}{EndSwitch}[1]{\algorithmicswitch\ #1\ \algorithmicdo}{\algorithmicend\ \algorithmicswitch}%
\algdef{SE}[CASE]{Case}{EndCase}[1]{\algorithmiccase\ #1}{\algorithmicend\ \algorithmiccase}%
\algtext*{EndSwitch}%
\algtext*{EndCase}%

\subsubsection{Experiment Setup} 
We evaluated the performance of the generator models, PPO, PCG-HR, and PCG-RD. We compiled 300 game results at each step to ensure an accurate evaluation of the generated contents.
The heuristic playtesting agent (PT-HR in Section \ref{sec:rob_exp_setup}) was employed during the playtesting process.
Additionally, this section focuses on demonstrating the generation of skill contents, and the heuristic method speeds up the experiment time and enhances the reliability of the evaluation.

Since PPO is trained from a randomly initialized neural network, we conduct the experiment five times to mitigate performance variations caused by initial weights.
After training the PPO models for 20K steps, we generate 1K skills for each model, resulting in a total of 5K samples (5 models $\times{}$ 1K samples = 5K samples).
All experimental results are summarized by averaging the simulated game results across the five seed runs.

\subsubsection{Evaluation Metrics}
To evaluate the contents generated by PPO, PCG-HR, and PCG-RD agents is performed using two commonly used metrics in PCG studies: (1) Controllability and (2) Diversity. A lack of controllability implies that the agent cannot generate contents that fulfill the game designer's requirements, while a lack of diversity results in repetitive content that can only be used once.

\textbf{Controllability}
Controllability is measured by evaluating the error between the designer's desired target win rate ($W_t$) and the simulated results obtained from the generated skills ($W_c$).
The error measurement for controllability is defined in Equation \ref{eq:Controllability}, where $N$ represents the number of samples.
\begin{align}
    \label{eq:Controllability}
    WinrateError = \sum_{i}^{N}|W_{i,t}-W_{i,c}|
\end{align}
$W_t$ denotes the target win rate desired by the designer, and $W_c$ represents the win rate of the generated skill.
The equation calculates the mean error of the win rates of skills relative to the target value.
The lower error indicates that the skills produced by the generator are more closely aligned with the designer's requirements.

\textbf{Diversity} Diversity measures the variety of styles in which the contents are generated while still satisfying the intended conditions.
A higher diversity enriches the gameplay experience and provides designers with a wider range of options.
The procedure for sampling skills is the same as in controllability; however, we apply a root mean squared error (RMSE) $<0.1$ filter to evaluate the diversity of the contents.
This filtering ensures that the evaluated diversity represents the skill parameter ranges satisfying the designer-defined win rates, thereby establishing the diversity metric as a more meaningful benchmark for usability.
Thus, diversity assesses the extent to which the generator offers diverse options to the designer based on the designer's desired conditions.
A higher value indicates that the generator has generated contents with diverse skill parameter ranges, resulting in various game experiences even with the same balanced results ($W_t$).
To simplify the interpretation of skill diversity, we employ principal component analysis (PCA) \cite{wold1987principal}, a dimension reduction technique, to represent the four parameters as a single value.

\subsection{Experimental Result \& Discussion}
\label{sec:pcg_exp_result_discuss}
\subsubsection{Controllability}
\begin{table}[!ht]
\centering
\caption{Descriptive statistics on the controllability of the generative models}
\label{tab:pcg_controllbility_small}
\begin{tabular}{cl|cc|cc}
\toprule
 &  & \multicolumn{2}{c}{Generated ($W_{c}$)} & \multicolumn{2}{c}{RMSE ($\|W_{t}-W_{c}\|$)} \\
 &  & \multicolumn{2}{c}{Mean {($\pm$SD)}} & \multicolumn{2}{c}{Mean {($\pm$SD)}} \\
\begin{tabular}[x]{@{}c@{}}Target\\($W_{t}$)\end{tabular} & Method &  &  &  &  \\
\midrule
\multirow[c]{3}{*}{0.1} & PCG-HR &\multicolumn{2}{c}{ \cellcolor[HTML]{6eaed1} \color{black} {0.107} ($\pm{}$0.065) } &\multicolumn{2}{c}{ \cellcolor[HTML]{408fc1} \color{black} \textbf{{0.044} ($\pm{}$0.049)} }  \\
 & PPO &\multicolumn{2}{c}{ \cellcolor[HTML]{90c4dd} \color{black} {0.149} ($\pm{}$0.242) } &\multicolumn{2}{c}{ \cellcolor[HTML]{95c6df} \color{black} {0.156} ($\pm{}$0.191) }  \\
 & PCG-RD &\multicolumn{2}{c}{ \cellcolor[HTML]{f7f6f6} \color{black} {0.354} ($\pm{}$0.389) } &\multicolumn{2}{c}{ \cellcolor[HTML]{f1f4f6} \color{black} {0.335} ($\pm{}$0.321) }  \\
\midrule
\multirow[c]{3}{*}{0.2} & PCG-HR &\multicolumn{2}{c}{ \cellcolor[HTML]{b5d7e8} \color{black} {0.206} ($\pm{}$0.098) } &\multicolumn{2}{c}{ \cellcolor[HTML]{559ec9} \color{black} \textbf{{0.074} ($\pm{}$0.065)} }  \\
 & PPO &\multicolumn{2}{c}{ \cellcolor[HTML]{d6e7f1} \color{black} {0.266} ($\pm{}$0.343) } &\multicolumn{2}{c}{ \cellcolor[HTML]{d6e7f1} \color{black} {0.268} ($\pm{}$0.225) }  \\
 & PCG-RD &\multicolumn{2}{c}{ \cellcolor[HTML]{f8f2ee} \color{black} {0.366} ($\pm{}$0.381) } &\multicolumn{2}{c}{ \cellcolor[HTML]{f0f3f5} \color{black} {0.333} ($\pm{}$0.249) }  \\
\midrule
\multirow[c]{3}{*}{0.3} & PCG-HR &\multicolumn{2}{c}{ \cellcolor[HTML]{e2edf3} \color{black} {0.297} ($\pm{}$0.108) } &\multicolumn{2}{c}{ \cellcolor[HTML]{58a0ca} \color{black} \textbf{{0.079} ($\pm{}$0.073)} }  \\
 & PPO &\multicolumn{2}{c}{ \cellcolor[HTML]{bad9e9} \color{black} {0.217} ($\pm{}$0.189) } &\multicolumn{2}{c}{ \cellcolor[HTML]{9dcae1} \color{black} {0.169} ($\pm{}$0.119) }  \\
 & PCG-RD &\multicolumn{2}{c}{ \cellcolor[HTML]{fbe3d6} \color{black} {0.418} ($\pm{}$0.382) } &\multicolumn{2}{c}{ \cellcolor[HTML]{f6f6f6} \color{black} {0.348} ($\pm{}$0.196) }  \\
\midrule
\multirow[c]{3}{*}{0.4} & PCG-HR &\multicolumn{2}{c}{ \cellcolor[HTML]{fae4d7} \color{black} {0.416} ($\pm{}$0.122) } &\multicolumn{2}{c}{ \cellcolor[HTML]{61a6cd} \color{black} \textbf{{0.089} ($\pm{}$0.085)} }  \\
 & PPO &\multicolumn{2}{c}{ \cellcolor[HTML]{f8efea} \color{black} {0.377} ($\pm{}$0.315) } &\multicolumn{2}{c}{ \cellcolor[HTML]{dbe9f1} \color{black} {0.276} ($\pm{}$0.153) }  \\
 & PCG-RD &\multicolumn{2}{c}{ \cellcolor[HTML]{f9ede7} \color{black} {0.381} ($\pm{}$0.394) } &\multicolumn{2}{c}{ \cellcolor[HTML]{f7f3f0} \color{black} {0.362} ($\pm{}$0.154) }  \\
\midrule
\multirow[c]{3}{*}{0.5} & PCG-HR &\multicolumn{2}{c}{ \cellcolor[HTML]{f9c5ab} \color{black} {0.489} ($\pm{}$0.133) } &\multicolumn{2}{c}{ \cellcolor[HTML]{68aacf} \color{black} \textbf{{0.099} ($\pm{}$0.089)} }  \\
 & PPO &\multicolumn{2}{c}{ \cellcolor[HTML]{fcdfce} \color{black} {0.435} ($\pm{}$0.334) } &\multicolumn{2}{c}{ \cellcolor[HTML]{e2edf3} \color{black} {0.296} ($\pm{}$0.167) }  \\
 & PCG-RD &\multicolumn{2}{c}{ \cellcolor[HTML]{f9ebe3} \color{black} {0.390} ($\pm{}$0.391) } &\multicolumn{2}{c}{ \cellcolor[HTML]{f8eee8} \color{black} {0.379} ($\pm{}$0.144) }  \\
\midrule
\multirow[c]{3}{*}{0.6} & PCG-HR &\multicolumn{2}{c}{ \cellcolor[HTML]{e6856a} \color{white} {0.592} ($\pm{}$0.132) } &\multicolumn{2}{c}{ \cellcolor[HTML]{68aacf} \color{black} \textbf{{0.097} ($\pm{}$0.090)} }  \\
 & PPO &\multicolumn{2}{c}{ \cellcolor[HTML]{e27d63} \color{white} {0.606} ($\pm{}$0.309) } &\multicolumn{2}{c}{ \cellcolor[HTML]{dbe9f1} \color{black} {0.276} ($\pm{}$0.138) }  \\
 & PCG-RD &\multicolumn{2}{c}{ \cellcolor[HTML]{fae7db} \color{black} {0.409} ($\pm{}$0.395) } &\multicolumn{2}{c}{ \cellcolor[HTML]{fae7db} \color{black} {0.405} ($\pm{}$0.168) }  \\
\midrule
\multirow[c]{3}{*}{0.7} & PCG-HR &\multicolumn{2}{c}{ \cellcolor[HTML]{c94641} \color{white} {0.684} ($\pm{}$0.113) } &\multicolumn{2}{c}{ \cellcolor[HTML]{5ea4cc} \color{black} \textbf{{0.088} ($\pm{}$0.073)} }  \\
 & PPO &\multicolumn{2}{c}{ \cellcolor[HTML]{d45d4b} \color{white} {0.652} ($\pm{}$0.354) } &\multicolumn{2}{c}{ \cellcolor[HTML]{e1ecf3} \color{black} {0.292} ($\pm{}$0.205) }  \\
 & PCG-RD &\multicolumn{2}{c}{ \cellcolor[HTML]{f9ece5} \color{black} {0.387} ($\pm{}$0.397) } &\multicolumn{2}{c}{ \cellcolor[HTML]{fcdcc8} \color{black} {0.445} ($\pm{}$0.239) }  \\
\midrule
\multirow[c]{3}{*}{Mean} & PCG-HR & \multicolumn{2}{c}{ \cellcolor[HTML]{ffffff} \color{black} \textbf{-} } &\multicolumn{2}{c}{ \cellcolor[HTML]{5ba2cb} \color{black} \textbf{{0.081} ($\pm{}$0.078)} }  \\
 & PPO & \multicolumn{2}{c}{ \cellcolor[HTML]{ffffff} \color{black} \textbf{-} } &\multicolumn{2}{c}{ \cellcolor[HTML]{cee3ef} \color{black} {0.248} ($\pm{}$0.183) }  \\
 & PCG-RD & \multicolumn{2}{c}{ \cellcolor[HTML]{ffffff} \color{black} \textbf{-} } &\multicolumn{2}{c}{ \cellcolor[HTML]{f8f0ec} \color{black} {0.373} ($\pm{}$0.221) }  \\
\bottomrule
\end{tabular}
\end{table}
The descriptive results for measuring the controllability of the generated win rate targets ($W_{t}$) are listed in Table \ref{tab:pcg_controllbility_small}. Cells with a blue background color indicate lower values, while red cells indicate higher values. We trained the PPO model using seven target parameters ranging from 0.1 to 0.7 and compared its performance with the mentioned approaches. Each of the three models generated $N=1,000$ skill parameters for each target ($W_{t}$) and was evaluated using the heuristic playtesting agent (PT-HR).

According to this criterion, the PCG-HR agent demonstrated the best performance, followed by the PPO agent in second place, and the PCG-RD agent in last place.
The PCG-HR agent consistently achieved the lowest RMSE values across all target win rate conditions. On average, its error in win rate estimation was approximately 8.1\%.
Surprisingly, the PPO agent performed worse than the heuristic agent. However, we observed that the PPO model was trained to minimize win rate errors compared to the random agent.
The suboptimal performance of the PPO agent can be attributed to the presence of noise in the reward signal, indicating that the win rate measurement could be improved by increasing the number of simulations.
The stochastic reward signal had a negative impact on the learning capabilities of the DRL-based models.

Despite its second-place ranking in this evaluation, the PPO agent has an advantage as a learning-based algorithm when it comes to skill sampling. Unlike the PCG-HR agent, which requires a simulation process before executing an action, the PPO agent can sample actions without relying on playtesting results. This feature reduces the computational cost associated with content generation. Consequently, the DRL-based method (PPO) can efficiently sample the skills if its performance is improved to match that of the heuristic agent.

\subsubsection{Diversity}
\begin{table}[!ht]
\centering
\caption{Descriptive statistics on the diversity of the generated contents}
\label{tab:pcg_diversity_small}
\begin{tabular}{cl|cccc|c}
\toprule
 &  & Range & \begin{tabular}[x]{@{}c@{}}Cool\\Time\end{tabular} & \begin{tabular}[x]{@{}c@{}}Cast\\Time\end{tabular} & Damage & PCA \\
 &  & {($\pm$SD)} & {($\pm$SD)} & {($\pm$SD)} & {($\pm$SD)} & {($\pm$SD)} \\
\begin{tabular}[x]{@{}c@{}}Target\\($W_{t}$)\end{tabular} & Method &  &  &  &  &  \\
\midrule
\multirow[c]{3}{*}{0.1} & PCG-HR & \cellcolor[HTML]{c6dbef} \color{black} 0.258 & \cellcolor[HTML]{feeede} \color{black} 0.300 & \cellcolor[HTML]{e9f6e5} \color{black} 0.269 & \cellcolor[HTML]{fee7dd} \color{black} 0.265 & \cellcolor[HTML]{f9f9f9} \color{black} 0.302 \\
 & PPO & \cellcolor[HTML]{a8cee4} \color{black} \textbf{0.315} & \cellcolor[HTML]{fda761} \color{black} \textbf{0.482} & \cellcolor[HTML]{ecf7e8} \color{black} 0.258 & \cellcolor[HTML]{fb8c6c} \color{black} \textbf{0.472} & \cellcolor[HTML]{b5b5b5} \color{black} \textbf{0.625} \\
 & PCG-RD & \cellcolor[HTML]{c3d9ee} \color{black} 0.266 & \cellcolor[HTML]{fee8d3} \color{black} 0.325 & \cellcolor[HTML]{e2f3dc} \color{black} \textbf{0.294} & \cellcolor[HTML]{fee3d6} \color{black} 0.283 & \cellcolor[HTML]{f5f5f5} \color{black} 0.336 \\
\midrule
\multirow[c]{3}{*}{0.2} & PCG-HR & \cellcolor[HTML]{c4daee} \color{black} 0.262 & \cellcolor[HTML]{fee8d3} \color{black} 0.324 & \cellcolor[HTML]{eef8eb} \color{black} 0.249 & \cellcolor[HTML]{fee1d4} \color{black} 0.288 & \cellcolor[HTML]{f9f9f9} \color{black} 0.298 \\
 & PPO & \cellcolor[HTML]{93c4de} \color{black} \textbf{0.351} & \cellcolor[HTML]{fdd7b0} \color{black} \textbf{0.381} & \cellcolor[HTML]{e3f4dd} \color{black} \textbf{0.292} & \cellcolor[HTML]{fca284} \color{black} \textbf{0.428} & \cellcolor[HTML]{e6e6e6} \color{black} \textbf{0.421} \\
 & PCG-RD & \cellcolor[HTML]{b9d5ea} \color{black} 0.284 & \cellcolor[HTML]{fee7d1} \color{black} 0.329 & \cellcolor[HTML]{e7f5e2} \color{black} 0.279 & \cellcolor[HTML]{fdd6c5} \color{black} 0.316 & \cellcolor[HTML]{f1f1f1} \color{black} 0.359 \\
\midrule
\multirow[c]{3}{*}{0.3} & PCG-HR & \cellcolor[HTML]{bfd8ec} \color{black} \textbf{0.271} & \cellcolor[HTML]{feebd8} \color{black} 0.312 & \cellcolor[HTML]{eaf7e6} \color{black} \textbf{0.265} & \cellcolor[HTML]{fee0d2} \color{black} 0.292 & \cellcolor[HTML]{f7f7f7} \color{black} 0.312 \\
 & PPO & \cellcolor[HTML]{d2e3f3} \color{black} 0.218 & \cellcolor[HTML]{fdd1a5} \color{black} \textbf{0.398} & \cellcolor[HTML]{ebf7e7} \color{black} 0.261 & \cellcolor[HTML]{fcc7b1} \color{black} \textbf{0.349} & \cellcolor[HTML]{d8d8d8} \color{black} \textbf{0.488} \\
 & PCG-RD & \cellcolor[HTML]{c1d9ed} \color{black} 0.268 & \cellcolor[HTML]{fee7d0} \color{black} 0.331 & \cellcolor[HTML]{ebf7e7} \color{black} 0.264 & \cellcolor[HTML]{fddaca} \color{black} 0.308 & \cellcolor[HTML]{f6f6f6} \color{black} 0.329 \\
\midrule
\multirow[c]{3}{*}{0.4} & PCG-HR & \cellcolor[HTML]{c8dcef} \color{black} 0.252 & \cellcolor[HTML]{feebd8} \color{black} 0.312 & \cellcolor[HTML]{f2faef} \color{black} 0.233 & \cellcolor[HTML]{fee4d8} \color{black} 0.278 & \cellcolor[HTML]{f9f9f9} \color{black} 0.304 \\
 & PPO & \cellcolor[HTML]{deebf7} \color{black} 0.180 & \cellcolor[HTML]{fdd8b3} \color{black} \textbf{0.377} & \cellcolor[HTML]{c5e8be} \color{black} \textbf{0.364} & \cellcolor[HTML]{fb8a6a} \color{black} \textbf{0.476} & \cellcolor[HTML]{c8c8c8} \color{black} \textbf{0.559} \\
 & PCG-RD & \cellcolor[HTML]{bcd7eb} \color{black} \textbf{0.277} & \cellcolor[HTML]{fee6cf} \color{black} 0.333 & \cellcolor[HTML]{e4f4df} \color{black} 0.287 & \cellcolor[HTML]{fee1d3} \color{black} 0.292 & \cellcolor[HTML]{f2f2f2} \color{black} 0.358 \\
\midrule
\multirow[c]{3}{*}{0.5} & PCG-HR & \cellcolor[HTML]{c6dbef} \color{black} 0.257 & \cellcolor[HTML]{feead6} \color{black} 0.316 & \cellcolor[HTML]{f1faef} \color{black} 0.236 & \cellcolor[HTML]{fee6db} \color{black} 0.269 & \cellcolor[HTML]{f8f8f8} \color{black} 0.310 \\
 & PPO & \cellcolor[HTML]{d7e6f4} \color{black} 0.204 & \cellcolor[HTML]{fdd3a8} \color{black} \textbf{0.393} & \cellcolor[HTML]{b7e2b0} \color{black} \textbf{0.391} & \cellcolor[HTML]{fcad91} \color{black} \textbf{0.405} & \cellcolor[HTML]{d1d1d1} \color{black} \textbf{0.519} \\
 & PCG-RD & \cellcolor[HTML]{bbd6eb} \color{black} \textbf{0.278} & \cellcolor[HTML]{fee8d2} \color{black} 0.328 & \cellcolor[HTML]{e6f5e1} \color{black} 0.280 & \cellcolor[HTML]{fee6db} \color{black} 0.269 & \cellcolor[HTML]{f6f6f6} \color{black} 0.326 \\
\midrule
\multirow[c]{3}{*}{0.6} & PCG-HR & \cellcolor[HTML]{c9ddf0} \color{black} 0.247 & \cellcolor[HTML]{feebd8} \color{black} 0.313 & \cellcolor[HTML]{f1faef} \color{black} 0.236 & \cellcolor[HTML]{fee6db} \color{black} 0.269 & \cellcolor[HTML]{f9f9f9} \color{black} 0.303 \\
 & PPO & \cellcolor[HTML]{f7fbff} \color{black} 0.105 & \cellcolor[HTML]{fef2e5} \color{black} 0.284 & \cellcolor[HTML]{f7fcf5} \color{black} 0.214 & \cellcolor[HTML]{fff5f0} \color{black} 0.212 & \cellcolor[HTML]{ffffff} \color{black} 0.258 \\
 & PCG-RD & \cellcolor[HTML]{c4daee} \color{black} \textbf{0.261} & \cellcolor[HTML]{feead6} \color{black} \textbf{0.317} & \cellcolor[HTML]{dff2d9} \color{black} \textbf{0.302} & \cellcolor[HTML]{fddbcb} \color{black} \textbf{0.306} & \cellcolor[HTML]{ededed} \color{black} \textbf{0.385} \\
\midrule
\multirow[c]{3}{*}{0.7} & PCG-HR & \cellcolor[HTML]{c6dbef} \color{black} 0.257 & \cellcolor[HTML]{fef0e1} \color{black} 0.293 & \cellcolor[HTML]{f4faf1} \color{black} 0.227 & \cellcolor[HTML]{fee8de} \color{black} 0.263 & \cellcolor[HTML]{fafafa} \color{black} 0.293 \\
 & PPO & \cellcolor[HTML]{dae8f5} \color{black} 0.194 & \cellcolor[HTML]{fff5eb} \color{black} 0.271 & \cellcolor[HTML]{98d493} \color{black} \textbf{0.447} & \cellcolor[HTML]{fcac90} \color{black} \textbf{0.406} & \cellcolor[HTML]{b5b5b5} \color{black} \textbf{0.625} \\
 & PCG-RD & \cellcolor[HTML]{c4daee} \color{black} \textbf{0.262} & \cellcolor[HTML]{fee9d5} \color{black} \textbf{0.322} & \cellcolor[HTML]{ecf7e8} \color{black} 0.258 & \cellcolor[HTML]{fddbcb} \color{black} 0.305 & \cellcolor[HTML]{f2f2f2} \color{black} 0.358 \\
\midrule
\multirow[c]{3}{*}{Mean} & PCG-HR & \cellcolor[HTML]{c6dbef} \color{black} 0.258 & \cellcolor[HTML]{feecd9} \color{black} 0.310 & \cellcolor[HTML]{eff9ec} \color{black} 0.246 & \cellcolor[HTML]{fee5d9} \color{black} 0.275 & \cellcolor[HTML]{f9f9f9} \color{black} 0.303 \\
 & PPO & \cellcolor[HTML]{d1e2f2} \color{black} 0.237 & \cellcolor[HTML]{fddbb9} \color{black} \textbf{0.375} & \cellcolor[HTML]{d8f0d2} \color{black} \textbf{0.328} & \cellcolor[HTML]{fcb297} \color{black} \textbf{0.402} & \cellcolor[HTML]{d6d6d6} \color{black} \textbf{0.513} \\
 & PCG-RD & \cellcolor[HTML]{bfd8ec} \color{black} \textbf{0.271} & \cellcolor[HTML]{fee8d3} \color{black} 0.326 & \cellcolor[HTML]{e6f5e1} \color{black} 0.281 & \cellcolor[HTML]{fdded0} \color{black} 0.297 & \cellcolor[HTML]{f3f3f3} \color{black} 0.351 \\
\bottomrule
\end{tabular}
\end{table}
Table \ref{tab:pcg_diversity_small} presents the descriptive results for measuring the diversity of the generated content characteristics. Different background colors represent each parameter type, with darker colors indicating higher values. We provide the standard deviation (SD) values for each skill parameter in separate columns of Table \ref{tab:pcg_diversity_small}. The SD value is calculated using $N=300$ samples for each parameter, and the samples are filtered based on a threshold (RMSE$<0.1$) to measure diversity only with valid skills that meet the designer’s requirements. A higher SD value indicates that the generator generated contents with diverse skill parameter ranges, resulting in various game experiences even with the same balanced results ($W_t$).
To better understand the diversity of the skills, which can be challenging with separated variance results, we employed the principal component analysis (PCA) \cite{abdi2010principal} technique, a dimension reduction technique, to represent the four parameters in a one-dimensional space.
By representing the skills in this space, we measured the variance of the skills using a single value that encompasses all the parameters of the skill content.
The average SD is calculated across all win rates.
The best values are marked in bold, and PPO outperformed the other models in diversity, showing the best performance in 6 out of 7 conditions.
This result implies that the PPO agent has a good ability to provide rich options for designers even when given a specific condition.

\subsubsection{Discussion}
\label{sec:pcg_discussion}
\begin{figure}[!h]
    \centering
    \includegraphics[width=1.0\linewidth]{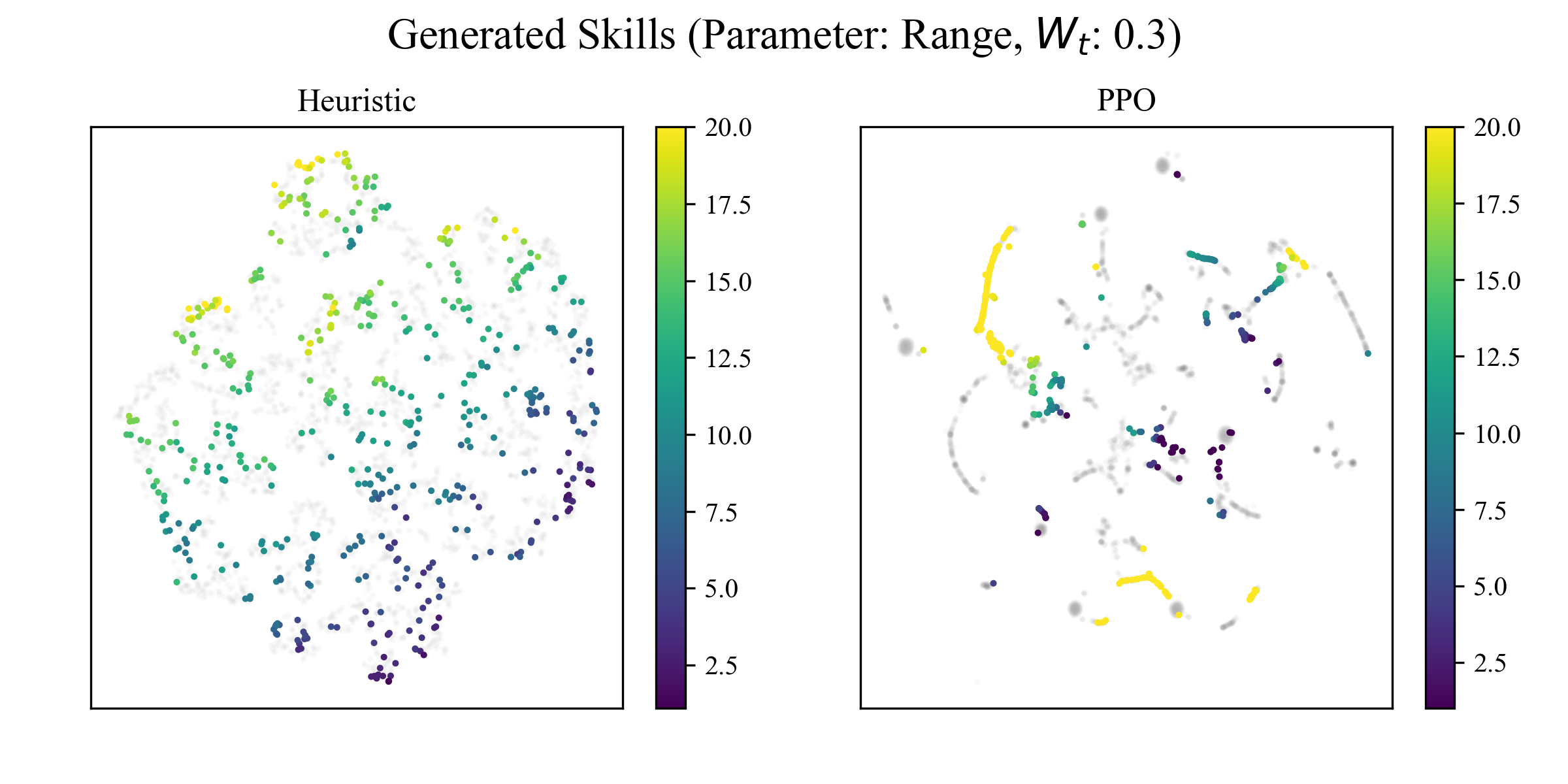}
    \caption{Visualization of generated skills in a two-dimensional plot using t-SNE, based on range parameter. Totally, 2,100 entries (300 samples $\times$ 7 target win rates) are involved, respectively. Note that gray dots denote all win rates, and colorized dots denote a specific target win rate (in this case, $W_t=0.3$).}
    \label{fig:range_visualization_compare}
\end{figure}

In this section, we summarize the two benchmark results, controllability, and diversity, and conduct further analysis.
The controllability results indicate that the heuristic agent outperformed other methods, including PPO, and random agents.
In the evaluation of diversity, we proposed a method to measure content diversity using variances of multiple parameters.
In diversity criteria, the PPO agent outperformed other methods, showing the highest variance in generated skill parameters. 
For further analysis, we utilized t-distributed stochastic neighbor embedding (t-SNE) \cite{van2008visualizing}, a popular dimensional reduction technique for data visualization. 
Figure \ref{fig:range_visualization_compare} illustrates examples of visualizations for the heuristic and PPO agents, respectively.

This study sheds light on the contrasting distribution patterns between the heuristic and PPO approaches in skill generation.
The heuristic approach exhibits a continuous distribution (Heuristic in Fig. \ref{fig:range_visualization_compare}), indicating a wide range of skills. In contrast, the PPO approach shows clustering (PPO in Fig. \ref{fig:range_visualization_compare}), suggesting a narrower focus on skill creation despite its distinct characteristics.
This implies that RL is capable of identifying and leveraging specific skill parameters to achieve optimal results.
However, as the RL aims to determine an optimal solution, its exploration of alternative states may be restricted.
Instead of seeking a single optimal solution, an alternative approach is needed to encourage the algorithm to discover different states and their corresponding skill parameters.
The skill parameters generated through PPO exhibit clustering compared to the PCG-HR approach. However, it can be observed that they are generally spread out, indicating higher degree of diversity in PPO which can also be verified numerically.

In this study, the heuristic algorithm was successfully employed by selecting and implementing four skill parameters, enabling experiments to be conducted at a usable level. However, commercial game environments possess complex variables that are challenging to hand-craft, necessitating a more comprehensive, and automated approach. Future research should focus on developing methodologies that allow RL to autonomously discover states and corresponding skill parameters in complex environments, rather than solely manipulating parameter values. Accordingly, the adaptability of the algorithm can be enhanced to a broader range of scenarios and its effectiveness can be improved for generating optimized skill parameters.

In conclusion, this study offers valuable insights into the distribution patterns and characteristics of heuristic and PPO approaches in skill parameter generation. The heuristic approach showcases versatility by generating a wide range of skills, while the PPO approach exhibits distinct characteristics albeit with a narrower focus. To overcome limitations and enhance the effectiveness of RL, future research should explore approaches that foster the exploration of diverse states, rather than solely pursuing a singular optimal solution.





\section{Conclusion \& Future Work}
In this paper, we introduce two benchmarks for training playtesting agents and generator models. We have developed a multiplayer game environment, specifically a boss raid benchmark, based on several MMORPG games. The customizable game components are designed to resemble the settings found in commercial-level games. The content generation track presents a novel approach to generating game skills, while the playtesting track emphasizes the importance of generalization ability when evaluating newly generated content. We anticipate that the development of the generate-and-test process will facilitate the integration of research-level algorithms into the industry. Furthermore, this environment offers game researchers the opportunity to explore new studies with highly customizable game variables.

Note that the current benchmarks in RaidEnv provide a simplistic setting. They are based on the assumption of a homogeneous agent, focused on target damage-based skills, and feature relatively simple content mechanisms, which limit the demonstration of more complex game transitions. Future endeavors should consider forming benchmarks that involve heterogeneous agents, such as various classes (tankers, proximity dealers, distance dealers), robustness to different content parameters, and content with more intricate mechanisms (bosses, skills, etc.), which would better align with the collaborative nature of the game industry.

Our framework offers numerous avenues for future work in multiagent systems and content generation, which were not explored in this study.
Firstly, we acknowledge the limitation of content types generated in this work, as it only focuses on skill contents as an example.
However, the environment encompasses a wide range of game contents, including player characteristics, and 15 game skills, which were not utilized in the experiment.
Increasing the controllable parameters may raise the complexity of action spaces, but it could enrich the benchmarks and provide valuable insights into controlling generative models to produce reliable outputs. Diversifying the generated contents, characters, and skills would create more diverse cooperative raid benchmarks. Secondly, training robust agents in heterogeneous settings remains a task for future exploration. In this work, we employed a homogeneous agent setting where the characteristics and skills are the same for all agents during playtesting.
However, in real boss raid scenarios, the performance of playtesting agents may decrease when their teammates vary. Therefore, the issue of generalization to different teammates becomes a significant challenge. Although we have utilized a limited set of features in our environment to maintain reasonable benchmark difficulty, we believe that this environment holds the potential to serve as a research platform that closely resembles commercial games.

\newpage

\section{Appendix}
\subsection{PCG \& Playtesting Heuristic Agent Algorithm}
\label{sec:pcg-hr algo}
Algorithm \ref{alg:Heuristic play-testing} showed the mechanism of playtesting heuristic (PT-HR) agent, and Algorithm \ref{alg:heuristic_pcg} showed the mechanism of PCG-heuristic (PCG-HR) agent.

\algrenewcommand\algorithmicrequire{\textbf{Input:}}
\algrenewcommand\algorithmicensure{\textbf{Output:}}
\begin{algorithm}[H]
\caption{Heuristic Playtesting Agent}\label{alg:Heuristic play-testing}
\begin{algorithmic}
\Require 
    Heuristic agent $\textit{H}$,
    Target agent $\textit{T}$, \\
    Heuristic agent skill $S_{\textit{H}}$,
    Heuristic agent position $P_{\textit{H}}$, \\
    Target agent position $P_{\textit{T}}$
\If {$ P_{\textit{T}} - P_{\textit{H}} < S_{\textit{H}}.range$} 
\Comment{Keep maximum skill range}
\State {Away from $\textit{T}$}
\Else
\State {Close to $\textit{T}$}
\EndIf
\If {$RandomNumber / 2 \equiv 0$}
\Comment{Move around target}
\State {Move Clockwise to $\textit{T}$}
\Else
\State {Move CounterClockwise to $\textit{T}$}
\EndIf

\If {$ P_{\textit{T}} - P_{\textit{H}} < S.range\ \land\ !{\textit{H}}.\text{Moving}\ \land\ !{\textit{H}}.\text{Casting}$}
\State Execute(skill)
\Comment{Attack target with skill}
\EndIf
\end{algorithmic}
\end{algorithm}
\begin{algorithm}[H]
\caption{Heuristic PCG Agent}\label{alg:heuristic_pcg}
\hspace*{\algorithmicindent} \textbf{Input:} Current skill parameters $S$  \\
\hspace*{\algorithmicindent} \hspace{1cm} Current winrate $W_c$, Target winrate $W_t$  \\
\begin{algorithmic}

     \State $S \gets$ $S\prime$
    \Comment{Copy the skill parameters}

    \State $selection \gets$ random value between 0 and 3
    
    \If {$W_c < W_t$}
        \Comment{Make the game easier}

        \Switch{$selection$}
            \Case{$0$}     \Comment{Increase the traveling range}
                \State{S.range $\gets$ S.range $+$ 1.5}
            \EndCase
            \Case{$1$}     \Comment{Decrease the cool time}
                \State{S.cast $\gets$ S.cooltime $-$ 10.0}
            \EndCase
            \Case{$2$}     \Comment{Decrease the cast time}
                \State{S.cost $\gets$ S.casttime $-$ 0.33}
            \EndCase
            \Case{$3$}     \Comment{Increase the attacking damage}
                \State{S.damage $\gets$ S.damage $+$ 0.33}
            \EndCase
        \EndSwitch
    \Else {}
        \Switch{$selection$}
            \Case{$0$}     \Comment{Increase the traveling range}
                \State{S.range $\gets$ S.range $-$ 1.5}
            \EndCase
            \Case{$1$}     \Comment{Decrease the cool time}
                \State{S.cast $\gets$ S.cooltime $+$ 10.0}
            \EndCase
            \Case{$2$}     \Comment{Decrease the cast time}
                \State{S.cost $\gets$ S.casttime $+$ 0.33}
            \EndCase
            \Case{$3$}     \Comment{Increase the attacking damage}
                \State{S.damage $\gets$ S.damage $-$ 0.33}
            \EndCase
        \EndSwitch
    \EndIf {}
    
\end{algorithmic}
\end{algorithm}

\newpage

\subsection{Playtesting Agent Result}
\begin{table*}[th]
\centering
\caption{Test and Train performance with adjustment of level difficulty}
\label{tab:robust_train_test_gap}
\begin{tabular}{@{}cc|cccccc@{}}
\toprule
 & Parameter & \multicolumn{5}{c}{$Range$} \\ \midrule
 & $\mathcal{S}$ & 3(Hard) & 4(Hard)& 5(Train) & 19(Easy) & 20(Easy) & Test Avg.\\ \midrule
 & 1 &\begin{tabular}[x]{@{}c@{}}0.004 ($\pm{0.005}$)\end{tabular} & 
 \begin{tabular}[x]{@{}c@{}}\textbf{0.553 ($\pm{0.085}$)}\end{tabular} & 
 \begin{tabular}[x]{@{}c@{}}\textbf{1.049 ($\pm{0.063}$)}\end{tabular} & 
 \begin{tabular}[x]{@{}c@{}}0.597 ($\pm{0.055}$)\end{tabular} &
 \begin{tabular}[x]{@{}c@{}}0.597 ($\pm{0.048}$)\end{tabular} &
 \begin{tabular}[x]{@{}c@{}}0.506 ($\pm{0.310}$)\end{tabular}\\
MA-POCA & 2 & \begin{tabular}[x]{@{}c@{}}\textbf{0.028 ($\pm{0.033}$)}\end{tabular} & 
\begin{tabular}[x]{@{}c@{}}\textbf{0.553 ($\pm{0.224}$)}\end{tabular}& 
\begin{tabular}[x]{@{}c@{}}1.004 ($\pm{0.122}$)\end{tabular}&
\begin{tabular}[x]{@{}c@{}}0.523 ($\pm{0.082}$)\end{tabular}&
\begin{tabular}[x]{@{}c@{}}0.525 ($\pm{0.079}$)\end{tabular}&
\begin{tabular}[x]{@{}c@{}}0.473 ($\pm{0.306}$)\end{tabular}\\
 & 3 & \begin{tabular}[x]{@{}c@{}}0.013 ($\pm{0.024}$)\end{tabular} &
 \begin{tabular}[x]{@{}c@{}}0.369 ($\pm{0.192}$)\end{tabular}& 
 \begin{tabular}[x]{@{}c@{}}0.630 ($\pm{0.268}$)\end{tabular}&
 \begin{tabular}[x]{@{}c@{}}0.826 ($\pm{0.109}$)\end{tabular} &
 \begin{tabular}[x]{@{}c@{}}0.778 ($\pm{0.151}$)\end{tabular}&
 \begin{tabular}[x]{@{}c@{}}\textbf{0.511 ($\pm{0.351}$)}\end{tabular}\\ 
  & 4 & \begin{tabular}[x]{@{}c@{}}0.000 ($\pm{0.000}$)\end{tabular} &
 \begin{tabular}[x]{@{}c@{}}0.016 ($\pm{0.015}$)\end{tabular}& 
 \begin{tabular}[x]{@{}c@{}}0.049 ($\pm{0.031}$)\end{tabular}&
 \begin{tabular}[x]{@{}c@{}}\textbf{0.980 ($\pm{0.017}$)}\end{tabular} &
 \begin{tabular}[x]{@{}c@{}}\textbf{0.971 ($\pm{0.022}$)}\end{tabular}&
 \begin{tabular}[x]{@{}c@{}}0.443 ($\pm{0.477}$)\end{tabular}\\ 
\hline
 \multicolumn{2}{c}{PT-Heuristic} & \begin{tabular}[x]{@{}c@{}}0.013 ($\pm{0.007}$)\end{tabular} &
 \begin{tabular}[x]{@{}c@{}}0.115 ($\pm{0.015}$)\end{tabular}& 
 \begin{tabular}[x]{@{}c@{}}0.199 ($\pm{0.038}$)\end{tabular}&
 \begin{tabular}[x]{@{}c@{}}0.89 ($\pm{0.011}$)\end{tabular} &
 \begin{tabular}[x]{@{}c@{}}0.868 ($\pm{0.021}$)\end{tabular}&
 \begin{tabular}[x]{@{}c@{}}0.441 ($\pm{0.396}$)\end{tabular}\\ \bottomrule
\end{tabular}%
\end{table*}

\begin{table*}[ht]

\center
\caption{The detailed description of the controllable parameters on RaidEnv}\label{tab:status_parameter}
\resizebox{\textwidth}{!}{
\begin{tabular}{ccccclllll}
\toprule 
\multicolumn{2}{c}{\textbf{Parameter}} & \textbf{Name} & \multicolumn{1}{c}{\textbf{Data Type}} & \multicolumn{1}{c}{\textbf{Range}} & \multicolumn{1}{c}{\textbf{Description}} \\ \midrule

\multirow{26}{*}{Character Statistics}
                           & \multirow{5}{*}{Status}    & Health Point    & integer  & [0, 1000]  & Maximum number that agent can be dealing with the damage  \\ \cmidrule{3-6} 
                           &                            & Manna Point    & integer  & [0,100]  & The resource that used when agent active a skill  \\ \cmidrule{3-6} 
                           &                            & Spell Power    & integer  & [0,  100]  & Reference number used when calculating the skill damage  \\ \cmidrule{3-6} 
                           &                            & Movement Speed    & float  & [1,  2]  & Move speed of the agent  \\ \cmidrule{2-6} 
                           & \multirow{4}{*}{Attack}    & Power    & integer  & [0,  100]  & Reference number used when calculating the melee attack damage  \\ \cmidrule{3-6} 
                           &                            & Range    & integer  & [1,  10]  & Maximum distance of the melee attack can be reached  \\ \cmidrule{3-6} 
                           &                            & Speed    & float  & [1,  2]  & Number of melee attacks can be activated per second  \\ \cmidrule{2-6} 
                           & \multirow{4}{*}{Defensive} & Armor    & integer  & [0,  100]  & Ratio that agent dealt reduced damage  \\ \cmidrule{3-6} 
                           &                            & Evasion  & integer  & [0,  100]  & Probability that agent can be dodge melee attack or skill  \\ \cmidrule{3-6} 
                           &                            & Parry    & float  & [0,  100]  & Probability that agent can block melee attack or skill  \\ \cmidrule{2-6} 
                           & \multirow{4}{*}{Primary}   & Strength    & integer  & [1,  100]  & Measure of physical power and proficiency. Affects agent melee attack damage, etc. \\ \cmidrule{3-6} 
                           &                            & Agility  & integer  & [1,  100]  & Measure of swiftness. Affects agent evasion, accuracy, etc. \\ \cmidrule{3-6} 
                           &                            & Intelligence    & integer  & [1,  100]  & Measure of mental power and magical prowess. Affect agent magical attack damage, spell effectiveness, etc. \\ \cmidrule{2-6}
                           & \multirow{5}{*}{Secondary} & Critical    & integer  & [0,  100]  & Chance that agent ability can more effect than normal activation  \\ \cmidrule{3-6} 
                           &                            & Haste  & integer  & [0,  100]  & Determine the agent's action swiftness as in casting, cool time, attack speed, etc   \\ \cmidrule{3-6} 
                           &                            & Versatility    & integer  & [0,  100]  & Increase the agent's attack and defensive abilities effect  \\ \cmidrule{3-6} 
                           &                            & Mastery    & integer  & [0,  100]  & Enhance some abilities effect to emphasize character class mechanism  \\ \midrule
\multirow{26}{*}{Skill}
                           & \multirow{12}{*}{Information} & Name    & string  &-   & Name given to each skill  \\ \cmidrule{3-6} 
                           &                            & Trigger Type       & enum  & [0,  1]  & Determine the skill is active skill or passive skill \\ \cmidrule{3-6} 
                           &                            & Magic School & enum  & [0,  9]  & Determine the visual effects of the skill(e.g. fire, frost etc)  \\ \cmidrule{3-6} 
                           &                            & Hit Type     & enum  & [0,  1]  & Determine the skill is melee attack or skill   \\ \cmidrule{3-6} 
                           &                            & Target Type     & enum  & [0,  2]  & Type of target which the skill can be used on (e.g. target, non-target, region)  \\ \cmidrule{3-6} 
                           &                            & Projectile Speed     & float  & [0,  50]  & Speed of the projectile that reaches to the target  \\ \cmidrule{3-6} 
                           &                            & Affect on Ally     & bool  & True/False  & Determine that a agent can activate a skill to ally  \\ \cmidrule{3-6} 
                           &                            & Affect on Enemy     & bool  & True/False  & Determine that a agent can activate a skill to enemy  \\ \cmidrule{2-6}
                           & \multirow{12}{*}{Condition} & Cool Time    & float  & [0,  60]  & Amount of time required to Activate the skill again  \\ \cmidrule{3-6} 
                           &                            & Cast Time       & float  & [0,  2]  & Amount of time required to execute the skill after activation  \\ \cmidrule{3-6} 
                           &                            & Cost & float  & [0,  100]  & Amount of Mana required to Activate the skill  \\ \cmidrule{3-6} 
                           &                            & Range     & float  & [1,  20]  & Maximum distance of the skill can be reached  \\ \cmidrule{3-6} 
                           &                            & Charge     & integer  & [1,  3]  & Number of skill can be activated multiple times regardless of the Cool time  \\ \cmidrule{3-6}
                           &                            & Cast on moving     & bool  & True/False  & Determine that an agent can activate a skill during the moving  \\ \cmidrule{3-6} 
                           &                            & Cast on Casting     & bool  & True/False  & Determine that an agent can activate a skill during casting another skill  \\ \cmidrule{3-6} 
                           &                            & Cast on Channeling     & bool  & True/False  & Determine that an agent can activate a skill during executing another skill  \\ \cmidrule{2-6}
                           & \multirow{1}{*}{Coefficient} & Value    & float  & [0,  2]  & Scaling factor that determines the damage of the skill  \\ \midrule
\end{tabular}
}
\end{table*}

Table \ref{tab:robust_train_test_gap} showed the performance on train and test content parameters. The results showed that MA-POCA4 and MA-POCA3 are not trained perfectly in trained settings(seen). Especially, MA-POCA4 has shown that failure to learn about the strategy when their skill has short distance. Overall, MA-POCA3 showed the best average performance in unseen environments. However, since all agents are failure to generalize in skill range 3, it indicates this benchmark is challenging.

\newpage

\subsection{Full Environment Description}
\label{sec:full_env_desc}
The RaidEnv platform offers extensive customization options for various game features, which can be controlled using different variables. Figure \ref{fig:player_class_architecture} provides an overview of all the game components, while Table \ref{tab:env_summary} summarizes the detailed descriptions, data types, and ranges associated with these components. The range values presented in the table were determined by the authors but can be adjusted by other researchers as needed.

\newpage


\bibliographystyle{IEEEtran}
\bibliography{references}

\end{document}